\documentclass[twoside,11pt]{article}

 \usepackage[preprint]{jmlr2e}

\usepackage{natbib,hyperref} 
\let\cite\citep

\usepackage{subfig}
\usepackage{amsmath}
\usepackage{dsfont,amssymb}
\newtheorem{comment}{Comment}
\usepackage{graphicx}
\usepackage{multirow}
\usepackage{color}
\usepackage{comment}
\usepackage{microtype}
\newcommand{\red}[1]{\textcolor{red}{#1}}

\newcommand{\R}{\ensuremath{\raisebox{\depth}{\rotatebox{180}{R}}}}
\newcommand{\framework}{\ensuremath{\mathsf{Justicia}}}
\newcommand{\frameworkenum}{\ensuremath{\mathsf{Justicia\_enum}}}
\newcommand{\frameworklearn}{\ensuremath{\mathsf{Justicia\_learn}}}
\newcommand{\frameworkcond}{\ensuremath{\mathsf{Justicia\_cond}}}
\newcommand{\nonsensitive}{\ensuremath{X}}
\newcommand{\sensitive}{\ensuremath{A}}
\newcommand{\bool}{\ensuremath{B}}
\newcommand{\alg}{\mathcal{M}}
\newcommand{\abs}[1]{\ensuremath{\mid #1 \mid}}
\usepackage{algorithm}
\usepackage[noend]{algpseudocode}
\usepackage{booktabs}

\usepackage{tikz}
\usetikzlibrary{bayesnet,arrows}
\usetikzlibrary{shapes,fit,calc,positioning}
\tikzset{box/.style={draw, diamond, thick, text centered, minimum height=0.5cm, minimum width=1cm, text width=0.9cm}}
\tikzset{line/.style={draw, thick, -latex'}}
\allowdisplaybreaks

\newtheorem{lemmarep}{Lemma} %for appendix
\newtheorem{theoremrep}[lemmarep]{Theorem} %for appendix
\newtheorem{corollaryrep}[lemmarep]{Corollary} %for appendix

\ShortHeadings{}{Ghosh, Basu, and Meel}
\firstpageno{1}

\begin{document}

\title{Justicia: A Stochastic SAT Approach to Formally\\ Verify Fairness}

\author{\name Bishwamittra Ghosh \\
       \addr School of Computing\\
       National University of Singapore\\
       Singapore
       \AND
       \name Debabrota Basu \\
       \addr Department of Computer Science and Engineering\\
       Chalmers University of Technology\\
       G\"oteborg, Sweden\\
       \\
       Scool, Inria Lille- Nord Europe\\France
    	\AND
	   	\name Kuldeep S. Meel  \\
	   	\addr School of Computing\\
	   	National University of Singapore\\
	   	Singapore}

%\editor{Kevin Murphy and Bernhard Sch{\"o}lkopf}

\maketitle

\begin{abstract}
	%The pervasive growth of Machine learning (ML) has brought consequential changes to our socioeconomic and personal lives. 
	As a technology ML is oblivious to societal good or bad, and thus, the field of fair machine learning has stepped up to propose multiple mathematical definitions, algorithms, and systems to ensure different notions of fairness in ML applications.
	Given the multitude of propositions, it has become imperative to formally verify the fairness metrics satisfied by different algorithms on different datasets.
	In this paper, we propose a \textit{stochastic satisfiability} (SSAT) framework, {\framework}, that formally verifies different fairness measures of supervised learning algorithms with respect to the underlying data distribution.
	We instantiate {\framework} on multiple classification and bias mitigation algorithms, and datasets to verify different fairness metrics, such as disparate impact, statistical parity, and equalized odds.
	{\framework} is scalable, accurate, and operates on non-Boolean and compound sensitive attributes unlike existing distribution-based verifiers, such as FairSquare and VeriFair.
	Being distribution-based by design, {\framework} is more robust than the verifiers, such as AIF360, that operate on specific test samples.
	We also theoretically bound the finite-sample error of the verified fairness measure.
%	
%	
%	As machine learning is becoming the pervasive technology of our time, it is used for high-stake decision making like recidivism, loan sanctions, personalised medicines. This motivates proposition of multiple definitions and algorithms to ensure fairness and mitigate algorithmic bias in recent time. As these definitions focus on different aspects of disparity in output of learning algorithms and different methods to mitigate them, it is important to measure \textit{fairness} of this methods formally. In this paper, we propose verifying fairness of a supervised learning algorithm as a \textit{stochastic satisfiability} (SSAT) problem. Our SSAT formulation allows us to efficiently verify different fairness metrics, such as independence and separation, for pre-, in-, and post-processing type fairness algorithms.
%	We provide a theoretical analysis to measure the error in verification due to limited number of samples. We empirically demonstrate effectiveness of our SSAT formulation to verify fairness of different algorithms on different datasets and for different metrics. Our method can measure fairness violation among compound sensitive variables, such as gender and race, while the previous verification methods, such as FairSquare, can verify only for a given sensitive variable. Our experiments also show that our method is robust than the empirical evaluation-based techniques to measure fairness.
\end{abstract}
%%This abstract is now long to help us to write the intro. We should concise it later.
\section{Introduction}
%\paragraph{Motivation: ML + Fairness}
Machine learning (ML) is becoming the omnipresent technology of our time. 
ML algorithms are being used for high-stake decisions like college admissions, crime recidivism, insurance, and loan decisions.
Thus, human lives are now pervasively influenced by data, ML, and their inherent bias.

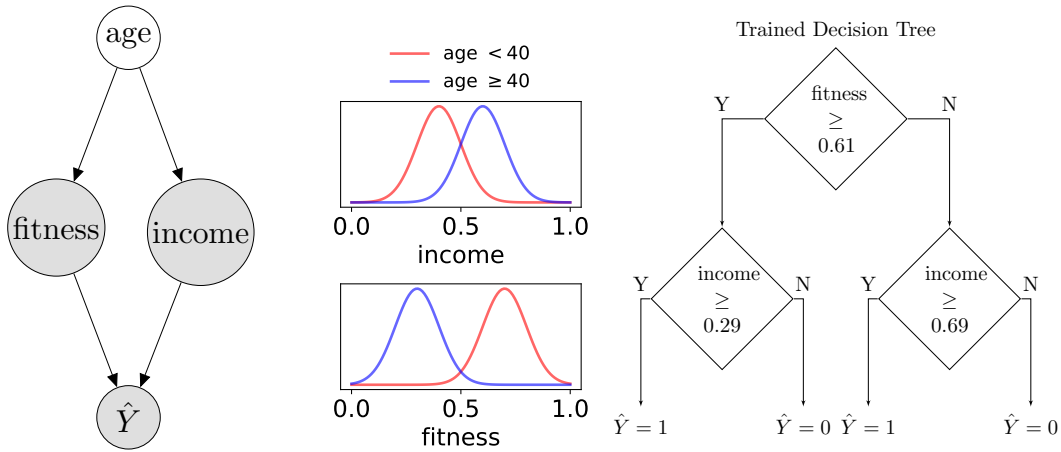
\begin{figure}[t!]%%\vspace*{-1em}
	\centering\hspace*{-4em}
	\begin{minipage}{.33\columnwidth}
		\centering
		\scalebox{0.8}{	
			\begin{tikzpicture}[x=1.5cm,y=1.8cm]
			% Define nodes
			\node[latent,scale=1.5] (a1) {$\textrm{age}$} ; %
			\node[obs, scale=1.5, below=of a1, xshift=-.8cm] (h) {$\textrm{fitness}$}; %
			\node[obs, scale=1.5, below=of a1, xshift=.8cm] (i) {$\textrm{income}$}; %
			\node[obs, scale=1.5, below=of h, xshift=.8cm] (p) {$\hat{Y}$}; %			
			%%add edge
			\edge[] {a1} {h,i} ;
			\edge[] {h,i} {p} ;
			\end{tikzpicture}
		}
	\end{minipage}\hspace*{-2em}
	\begin{minipage}{.33\columnwidth}
		\centering
		\includegraphics[scale=0.5]{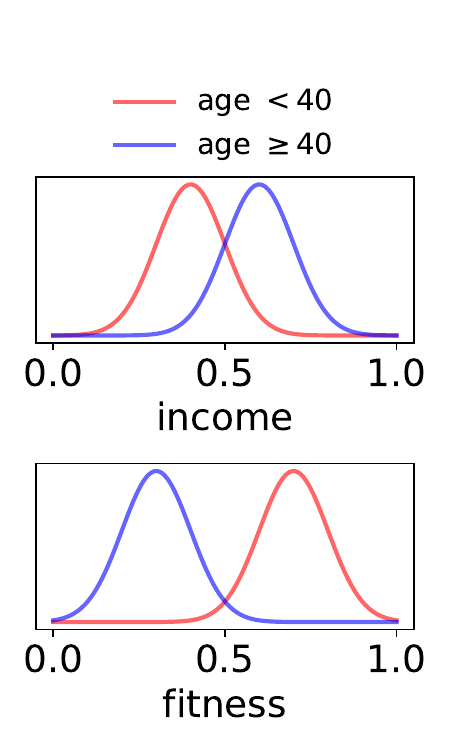}
	\end{minipage}\hspace*{-2em}
	\begin{minipage}{.33\columnwidth}
		\scalebox{0.48}{	
			\begin{tikzpicture}[x=1cm,y=1.8cm]
			\node [box, scale=1.5]                                    (p)      {fitness $\geq 0.61$};
			\node [scale=1.5, above= 0.1cm of p]  (a)    {Trained Decision Tree};
			\node [scale=1.5, box, below= of p, xshift=-2.1cm]    (a1)    {income\\ $\geq 0.29$};
			\node [scale=1.5, box, below= of p, xshift=2.1cm]     (a2)    {income $\geq 0.69$};
			\node [scale=1.5,below= of a1, xshift=-1.5cm]  (a11)    { $\hat{Y}= 1$};
			\node [scale=1.5,below= of a1, xshift=1.5cm]   (a12)    { $\hat{Y}=0 $};
			\node [scale=1.5,below= of a2, xshift=-1.5cm]  (a21)    { $\hat{Y}= 1$};
			\node [scale=1.5,below= of a2, xshift=1.5cm]  (a22)    { $\hat{Y}= 0$};
			\path [line] (p) -|         (a1) node [scale=1.5,midway, above]  {Y};
			\path [line] (p) -|         (a2) node [scale=1.5,midway, above]  {N};
			\path [line] (a1) -|       (a11) node [scale=1.5,midway, above]  {Y};
			\path [line] (a1) -|       (a12) node [scale=1.5,midway, above]  {N};
			\path [line] (a2) -|       (a21) node [scale=1.5,midway, above]  {Y};
			\path [line] (a2) -|       (a22) node [scale=1.5,midway, above]  {N};
			\end{tikzpicture}}
	\end{minipage}%
	\caption{A trained decision tree to learn eligibility for health insurance using age-dependent fitness and income indicators.}\label{fig:fair_example}%\vspace*{-2em}
\end{figure}

\begin{example}\label{example:intro}
Let us consider an example (Figure~\ref{fig:fair_example}) of deciding eligibility for health insurance depending on the fitness and income of the individuals of different age groups (20-40 and 40-60).
Typically, incomes of individuals increase as their ages increase while their fitness deteriorates.
We assume relation of income and fitness depends on the age as per the Normal distributions in Figure~\ref{fig:fair_example}.
Now, if we train a decision tree~\cite{narodytska2018learning} on these fitness and income indicators to decide the eligibility of an individual to get a health insurance, we observe that the `optimal' decision tree (ref. Figure~\ref{fig:fair_example}) selects a person above and below $40$ years with probabilities $0.18$ and $0.72$ respectively.
This simple example demonstrates that even if an ML algorithm does not explicitly learn to differentiate on the basis of a sensitive attribute, it discriminates different age groups due to the utilitarian sense of accuracy that it tries to optimize.
\end{example}

\paragraph{Fair ML.} Statistical discriminations caused by ML algorithms have motivated researchers to develop several frameworks to ensure fairness and several algorithms to mitigate bias.
Existing fairness metrics mostly belong to three categories: \textit{independence}, \textit{separation}, and \textit{sufficiency}~\cite{mehrabi2019survey}. 
Independence metrics, such as demographic parity, statistical parity, and group parity, try and ensure the outcomes of an algorithm to be independent of the groups that the individuals belong to~\cite{feldman2015certifying,dwork2012fairness}.
Separation metrics, such as equalized odds, define an algorithm to be fair if the probability of getting the same outcomes for different groups are same~\cite{hardt2016equality}.
Sufficiency metrics, such as counterfactual fairness, constrain the probability of outcomes to be independent of individual's sensitive data given their identical non-sensitive data~\cite{kusner2017counterfactual}. %,wu2019counterfactual

In Figure~\ref{fig:fair_example}, independence is satisfied if the probability of getting insurance is same for both the age groups. Separation is satisfied if the number of `actually' (ground-truth) ineligible and eligible people getting the insurance are same. Sufficiency is satisfied if the eligibility is independent of their age given their attributes are the same.
Thus, we see that the metrics of fairness can be contradictory and complimentary depending on the application and the data~\cite{corbett2018measure}.
Different algorithms have also been devised to ensure one or multiple of the fairness definitions.
These algorithms try to rectify and mitigate the bias in the data and thus in the prediction-model in three ways: \textit{pre-processing} the data~\cite{kamiran2012data,zemel2013learning,calmon2017optimized}, \textit{in-processing} the algorithm~\cite{zhang2018mitigating}, and \textit{post-processing} the outcomes~\cite{kamiran2012decision,hardt2016equality}.

\paragraph{Fairness Verifiers.} Due to the abundance of fairness metrics and difference in algorithms to achieve them, it has become necessary to verify different fairness metrics over datasets and algorithms. 

In order to verify fairness as a model property on a dataset, verifiers like \textit{FairSquare}~\cite{albarghouthi2017fairsquare} and \textit{VeriFair}~\cite{bastani2019probabilistic} have been proposed. 
%FairSquare verifies demographic parity and individual fairness as a numerical integration problem for a specific program semantics.
%VeriFair translates fairness metrics to an enumeration problem of a specified Boolean syntax.
%These papers operate for a specific Boolean sensitive attribute.
These verifiers are referred to as {\em distributional verifiers} owing to the fact that their inputs are a probability  distribution of the attributes in the dataset and a model of a suitable form, and their objective is to verify fairness w.r.t. the distribution and the model.
Though FairSquare and VeriFair are robust and have asymptotic convergence guarantees, we observe that they scale up poorly with the size of inputs and also do not generalize to non-Boolean and compound sensitive attributes.
In contrast to the distributional verifiers, another line of work, referred to as sample-based verifiers, has focused on the design of testing methodologies  on a given fixed data sample~\cite{galhotra2017fairness,aif360-oct-2018}. 
Since sample-based verifiers are dataset-specific, they generally do not provide robustness over the distribution.

%\blue{Other papers: Probabilistic Verification of Fairness Properties via	Concentration, Verifying Individual Fairness in Machine Learning Models~\cite{john2020verifying}}
Thus, a \textit{unified formal framework} to verify \textit{different fairness metrics} of an ML algorithm, which is \textit{scalable}, capable of \textit{handling compound protected groups}, \textit{robust} with respect to the test data, and \textit{operational on real-life} datasets and fairness-enhancing algorithms, is missing in the literature.

\paragraph{Our Contribution.} From this vantage point, \textit{we propose to model verifying different fairness metrics as a Stochastic Boolean Satisfiability (SSAT) problem}~\cite{littman2001stochastic}.
SSAT was originally introduced by ~\cite{papadimitriou1985games} to model {\em games against nature}. In this work, we primarily focus on reductions to the exist-random quantified fragment of SSAT, which is also known as E-MAJSAT~\cite{littman2001stochastic}. 
SSAT is a conceptual framework that has been employed to capture several fundamental problems in AI such as computation of maximum a posteriori (MAP) hypothesis~\cite{fremont2017maximum},  propositional probabilistic planning~\cite{majercik2007appssat},  and circuit verification~\cite{lee2018towards}.
Furthermore, our choice of SSAT as a target formulation is motivated by the recent algorithmic progress that has yielded efficient SSAT tools~\cite{lee2017solving,lee2018solving}.

%We formulate SSAT encodings of the fairness verification problems and two methods to evaluate them in order to verify independence and separation metrics for any supervised learning algorithm using a unified scheme.
%Our encodings not only allow us to compute for non-Boolean and compound sensitive attributes but also to scale significantly better than existing formal verifiers. 
%We perform experimental analysis for multiple fairness metrics, datasets, and algorithms to instantiate the efficiency and effectiveness of our approach.

Our contributions are summarised below:
\begin{itemize}
	\item We propose a unified SSAT-based approach, {\framework}, to verify independence and separation metrics of fairness for different datasets and classification algorithms.
	%\item \blue{{\framework} measures fairness metrics for pre-, in-, and post-processing algorithms with respect to the data  generating distribution}.
	\item Unlike previously proposed formal distributional verifiers, namely FairSquare and VeriFair, {\framework} verifies fairness for compound and non-Boolean sensitive attributes.%, and also \red{separation metrics}.
	\item Our experiments validate that our method is more accurate and scalable than the distributional verifiers, such as FairSquare and VeriFair, and more robust than the sample-based empirical verifiers, such as AIF360.
	\item We prove a finite-sample error bound on our estimated fairness metrics which is stronger than the existing asymptotic guarantees.
\end{itemize}

It is worth remarking that significant advances in AI bear testimony to the right choice of formulation, for example, formulation of planning as SAT~\cite{kautz1992planning}. In this context, we view that formulation of fairness as SSAT has potential to spur future work from both the modeling and encoding perspective as well as core algorithmic improvements in the underlying SSAT solvers.  

\iffalse
With this growing set of algorithms and definitions, it has become important to measure and verify fairness and bias of different algorithms and datasets. 
One popular approach is to use specific test dataset to compute the related statistical quantities and to certify fairness for that specific test dataset.
AIF360~\cite{aif360-oct-2018} provides a unified framework to implement multiple algorithms and to measure their fairness depending on such test datasets.
Though this method of verification works for a specified datasets, such verifiers do not explain how much a fairness measure depends on which sensitive attribute and is not robust to the selection of test dataset and its size.
\fi
\section{Background: Fairness and SSAT}
\label{sec:preliminaries}
In Section~\ref{sec:fairness}, we define different fairness metrics for a supervised learning problem. Following that, we discuss Stochastic Boolean Satisfiability (SSAT) problem in Section~\ref{sec:ssat}.

\subsection{Fairness Metrics for Machine Learning}\label{sec:fairness}
Let us represent a dataset $D$ as a collection of triples $(X, A, Y)$ sampled from an underlying data generating distribution $\mathcal{D} $.
$X \triangleq \lbrace \nonsensitive_1, \ldots, \nonsensitive_m\rbrace \in \mathbb{R}^m $ is the set of non-protected (or non-sensitive) attributes.
$A \triangleq \lbrace A_1, \ldots, A_n\rbrace$ is the set of categorical protected attributes.
$Y$ is the binary label (or class) of $(X, A)$. 
A compound protected attribute $\mathbf{a} = \{a_1, \ldots, a_n\}$ is a valuation to all $A_i$'s and represents a \textit{compound protected} group. For example, $A = \{\textrm{race}, \textrm{sex}\}$, where $\textrm{race} \in \{\textrm{Asian}, \textrm{Colour}, \textrm{White}\}$ and  $\textrm{sex} \in \{\textrm{female},  \textrm{male}\}$. 
Thus, $\mathbf{a} =  \{\textrm{Colour}, \textrm{female}\}$ is a compound protected group. 
We define $\alg \triangleq \Pr(\hat{Y}|X, A)$ to be a binary classifier trained from samples in the distribution $\mathcal{D} $. 
Here, $\hat{Y}$ is the predicted label (or class) of the corresponding data.

As we illustrated in Example~\ref{example:intro}, a classifier $\alg$ that solely optimizes accuracy, i.e., the average number of times $\hat{Y} = Y$, may discriminate certain compound protected groups over others~\cite{chouldechova2020snapshot}.
Now, we describe two family of fairness metrics that compute bias induced by a classifier and are later verified by {\framework}.

\subsubsection{Independence Metrics of Fairness.} 
The \textit{independence (or calibration) metrics} of fairness state that the output of the classifier should be independent of the compound protected group.
A notion of independence is referred to \textit{group fairness} that specifies an \textit{equal positive predictive value (PPV) across all compound protected groups} for an algorithm $\alg$, i.e., $\Pr[\hat{Y} = 1 | A = \mathbf{a}, \alg] = \Pr [\hat{Y} =1 | A = \mathbf{b}, \alg], \forall \mathbf{a}, \mathbf{b} \in A$.
Since satisfying group fairness exactly is hard, relaxations of group fairness, such as \textit{disparate impact} and \textit{statistical parity}~\cite{dwork2012fairness,feldman2015certifying}, are proposed. 

\textit{Disparate impact} (DI)~\cite{feldman2015certifying} measures the ratio of PPVs between the most favored group and least favored group, and prescribe it to be close to $1$. Formally, a classifier satisfies $(1 - \epsilon)$-disparate impact if, for $\epsilon \in [0,1] $,
\[
	\min_{\mathbf{a} \in A} \Pr[\hat{Y} =1 | \mathbf{a}, \alg]  \ge (1 - \epsilon) \max_{\mathbf{b} \in A} \Pr[\hat{Y} =1 | \mathbf{b}, \alg].
\]
Another popular relaxation of group fairness, \textit{statistical parity} (SP) measures the difference of PPV among the compound groups, and prescribe this to be near zero. Formally, an algorithm satisfies $\epsilon$-statistical parity if, for $\epsilon \in [0,1] $, 
\[
\max_{\mathbf{a}, \mathbf{b} \in A}|\Pr[\hat{Y} =1 | \mathbf{a}, \alg] - \Pr [\hat{Y} = 1| \mathbf{b}, \alg]| \le \epsilon.
\]
For both disparate impact and statistical parity, lower value of $\epsilon$ indicates higher group fairness of the classifier $\alg$.

\begin{comment}These two metrics of fairness has a disadvantage that a fully accurate classifier with different base rates (i.e., the proportion of actual positive prediction) in different protected groups may be considered unfair. As a result, similar individuals from different protected groups may be treated differently in order to achieve group fairness{\textemdash} such treatment is prohibited by law in some cases. 
\end{comment}

\subsubsection{Separation Metrics of Fairness.}
In the \textit{separation (or classification parity)} notion of fairness, the predicted label $\hat{Y}$ of a classifier $\alg$ is independent of the sensitive attributes $A$ given the actual class labels $Y$. In case of binary classifiers, a popular separation metric is \textit{equalized odds} (EO)~\cite{hardt2016equality} that computes the difference of false positive rates (FPR) and the difference of true positive rates (TPR) among all compound protected groups. 
Lower value of equalized odds indicates better fairness.
A classifier $\alg$ satisfies $\epsilon$-equalized odds if, for all compound protected groups $\mathbf{a}, \mathbf{b} \in A$,
\begin{equation}
\begin{split}
|\Pr[\hat{Y} =1 | A = \mathbf{a}, Y= 0  ] - \Pr [\hat{Y} = 1| A = \mathbf{b}, Y = 0]| &\le \epsilon, \\ 
|\Pr[\hat{Y} =1 | A = \mathbf{a}, Y= 1  ] - \Pr [\hat{Y} = 1| A = \mathbf{b}, Y = 1]| &\le \epsilon.\notag
\end{split}
\end{equation}

%\begin{comment}
%In contrast to disparate impact and statistical parity difference metrics, a fully accurate classifier will necessarily satisfy the two constraints in equalized odds measure. 

%\textbf{Group fairness.}  Group fairness requires the probability for an individual to be assigned the favorable outcome to be equal across the privileged  and unprivileged groups. Throughout the manuscript, by favorable outcome we refer to the predictor $ \hat{y} = 1 $ in  the standard binary classification setting. Now group fairness is formally defined in the following. 
%
%\[
%\Pr(\hat{y} =1 | a =1  ) = \Pr (\hat{y} = 1| a = 0)
%\]
%
%By abusing notations, we use the following shorthand. 
%
%\[
%\Pr_1(\hat{y} =1) = \Pr_0(\hat{y} = 1)
%\]
%
%
%
%
%In practice, the above definition can be relaxed. Thus we define the notion of $ \epsilon $-\textit{group fairness} in the following. 
%\[
% |\Pr_1(\hat{y} =1) - \Pr_0(\hat{y} = 1) |< \epsilon
%\]
%
%\end{comment}

In this paper, we formulate verifying the aforementioned independence and separation metrics of fairness as stochastic Boolean satisfiability (SSAT) problem, which we define next. 

\subsection{Stochastic Boolean Satisfiability (SSAT)}\label{sec:ssat}
Let $\mathbf{B}  = \{\bool_1, \dots, \bool_m\}  $ be a set of Boolean variables. A \textit{literal} is a variable $ \bool_i $ or its complement $ \neg \bool_i $. 
A propositional formula $\phi$ defined over $\mathbf{B}$ is in \textit{Conjunctive Normal Form (CNF)} if $\phi$   is  a conjunction of clauses and each clause is a disjunction of literals. 
%\red{DNF (disjunctive normal form) is the  complement of CNF where  the formula is a disjunction of clauses and each clause is  a conjunction of literals.} 
Let $ \sigma $ be an assignment to the  variables $ \bool_i \in \mathbf{B} $  such that $ \sigma(\bool_i) \in \{1, 0\} $ where $ 1 $ is logical TRUE and $ 0 $ is logical FALSE. The propositional  \textit{satisfiability} problem (SAT)~\cite{biere2009handbook} finds an assignment $ \sigma $ to all $ \bool_i \in  \mathbf{B} $ such that the formula $ \phi $ is evaluated to be $1$. 
In contrast to the SAT problem, the \textit{Stochastic Boolean Satisfiability} (SSAT) problem~\cite{littman2001stochastic} is concerned with the  probability of the satisfaction of the formula $\phi$. 
An SSAT formula is of the form
\begin{equation}\label{eq:ssat}
\Phi = Q_1\bool_1, \dots, Q_m \bool_m,\; \phi, 
\end{equation}
where $ Q_i \in \{\exists, \forall, \R^{p_i}\} $ is either of the existential ($\exists$), universal ($\forall$), or randomized ($\R^{p_i}$) quantifiers over the Boolean variable $\bool_i$ and $\phi$ is a quantifier-free CNF formula. In the SSAT formula $ \Phi $, the quantifier part $ Q_1\bool_1, \dots, Q_m \bool_m $ is known as the \textit{prefix} of the formula $ \phi $. In case of randomized quantification $ \R^{p_i} $, $ p_i \in [0,1] $ is the probability of $ \bool_i $ being assigned to $ 1 $. Given an SSAT formula $ \Phi $, let $ \bool $ be the outermost variable in the prefix. The satisfying probability of $ \Phi $ can be computed by the following \textit{rules}:
\begin{enumerate}
	\item $ \Pr[\text{TRUE}] = 1 $,  $ \Pr[\text{FALSE}] = 0 $, 
	\item $ \Pr [\Phi] = \max_{\bool} \{\Pr[\Phi|_{\bool}], \Pr[\Phi|_{\neg \bool}]\}$ if $ \bool $ is existentially quantified ($ \exists $), 
	\item $ \Pr [\Phi] = \min_{\bool} \{\Pr[\Phi|_{\bool}], \Pr[\Phi|_{\neg \bool}]\} $ if $ \bool $ is universally quantified ($ \forall $), 
	\item $ \Pr [\Phi] = p\Pr[\Phi|_{\bool}] + (1-p) \Pr[\Phi|_{\neg \bool}] $ if $ \bool $ is randomized quantified ($\R^{p}$) with probability $p$ of being $\text{TRUE}$,
\end{enumerate}
where $ \Phi|_{\bool} $ and $ \Phi|_{\neg \bool} $ denote the SSAT formulas derived by eliminating the outermost quantifier of $ \bool $  by substituting the value of $ \bool $ in the formula $ \phi $ with $ 1 $ and $ 0 $ respectively. In this paper, we focus on two specific types of SSAT formulas:  \textit{random-exist} (RE) SSAT and \textit{exist-random} (ER) SSAT. In the ER-SSAT (resp.\ RE-SSAT) formula, all existentially (resp.\ randomized) quantified variables are followed by randomized (resp.\ existentially) quantified variables in the prefix.

\textbf{Remark.} ER-SSAT problem is $\mathrm{NP}^{\mathrm{PP}}$-hard whereas RE-SSAT problem is $\mathrm{PP}^{\mathrm{NP}}$-complete~\cite{littman2001stochastic}.

The problem of SSAT and its variants have been pursued by theoreticians and practitioners alike for over three decades~\cite{majercik2005dc,fremont2017maximum,huang2006combining}. We refer the reader to~\cite{lee2017solving,lee2018solving} for detailed survey. It is worth remarking that the past decade has witnessed a significant performance improvements thanks to close integration of techniques from SAT solving with advances in weighted model counting~\cite{sang2004combining,chakraborty2013scalable,chakraborty2014distribution}.

\section{{\framework}: An SSAT Framework to Verify Fairness Metrics}
\label{sec:framework}
In this section, we present the primary contribution of this paper, {\framework}, which is an SSAT-based framework for verifying independence and separation metrics of fairness. 

Given a binary classifier $\alg$ and a probability distribution over dataset $(X,A,Y) \sim \mathcal{D} $, our goal is to verify whether $\alg$ achieves independence and separation metrics with respect to the distribution $\mathcal{D}$. We  focus on a classifier that can be translated to a CNF formula of Boolean variables $\mathbf{B} $. 
The probability $ p_i $ of $\bool_i \in \mathbf{B}$ being assigned to $1$ is induced by the data generating distribution $\mathcal{D}$. 
In order to verify fairness metrics in compound protected groups, we discuss an enumeration-based approach in Section~\ref{sec:enumeration_ssat} and an equivalent learning-based approach in Section~\ref{sec:learn_ssat}. 
%We conclude this section by proposing a conditional distribution based enumeration for compound protected groups in Section~\ref{sec:cond_ssat}. 
We conclude this section with a theoretical analysis for a high-probability error bound on the fairness metric in Section~\ref{sec:theory}. 

\iffalse
In this section, we present the main contribution of this paper, {\framework}, which is an SSAT framework for verifying independence and separation metrics of fairness. 
We first state the problem formally in Section~\ref{sec:problem_statement}. 
To verify fairness metrics in compound protected groups, we discuss an enumeration approach in Section~\ref{sec:enumeration_ssat} and an equivalent but more efficient learning approach in Section~\ref{sec:learn_ssat}. 
We conclude this section by proposing a conditional distribution based enumeration for compound protected groups in Section~\ref{sec:cond_ssat}.

\subsection{Problem Statement}
\label{sec:problem_statement}
Given a binary classifier $\alg$ and a probability distribution over dataset $(X,A,Y) \sim \mathcal{D} $, our goal is to verify whether $\alg$ achieves independence and separation metrics with respect to the distribution $\mathcal{D}$. We  focus on a classifier that can be translated to a CNF formula of Boolean variables $\mathbf{B} $. 
The probability $ p_i $ of $\bool_i \in \mathbf{B}$ being assigned to $1$ is induced by the data generating distribution $\mathcal{D}$. 
In our contribution, we reduce the verification problem to solving appropriately designed SSAT instances.
\fi 

\subsection{Evaluating Fairness with RE-SSAT Encoding}
\label{sec:enumeration_ssat}
In order to verify independence and separation metrics, the core component of {\framework} is to compute the positive predictive value $\Pr[\hat{Y} = 1 | A = \mathbf{a}]$ for a compound protected group $\mathbf{a}$.  For simplicity, we  initially make some assumptions and discuss their practical relaxations in Section~\ref{sec:practical-setting}.   
We first assume the classifier $\alg$ is representable as a CNF formula, namely $\phi_{\hat{Y}}$, such that $ \hat{Y} = 1 $ when $ \phi_{\hat{Y}}$ is satisfied and  $\hat{Y} =0$ otherwise. Since a Boolean CNF classifier is defined over Boolean variables, we  assume all attributes in $X$ and $A$ to be Boolean. Finally, we assume independence of non-protected attributes on protected attributes and $p_i $ is the  probability of the attribute $ \nonsensitive_i $ being assigned to $ 1 $ for any  $\nonsensitive_i \in X  $. 

Now, we define an RE-SSAT formula $\Phi_{\mathbf{a}}$ to compute the probability $\Pr[\hat{Y} = 1 | A = \mathbf{a}]$. In the prefix of $ \Phi_{\mathbf{a}} $,  all non-protected Boolean attributes in $X$ are assigned randomized quantification and they are followed by the protected Boolean attributes in $ A $ with existential quantification. The CNF formula $ \phi $ in $ \Phi_{\mathbf{a}} $ is constructed such that $ \phi $ encodes the event inside the target probability $ \Pr[\hat{Y} = 1 | A = \mathbf{a}] $. In order to encode the conditional $ A = \mathbf{a} $, we take the conjunction of the Boolean variables in $ A $ that symbolically specifies the compound protected group $ \mathbf{a} $. For example, we represent two protected attributes: race $ \in $ \{White, Colour\} and sex $ \in $ \{male, female\} by the Boolean variables $ R $ and $ S $  respectively. Thus, the compound groups $\{\textrm{White}, \textrm{male}\}$ and $\{\textrm{Colour}, \textrm{female}\}$ are represented by $ R \wedge S $ and $ \neg R \wedge \neg S $, respectively. Thus, the RE-SSAT formula for computing the probability  $ \Pr[\hat{Y} = 1 | A = \mathbf{a}] $ is
\begin{equation}	\label{eq:re}
\begin{split}
	\Phi_{\mathbf{a}} := \underbrace{\R^{p_{1}}\nonsensitive_1, \dots, \R^{p_{m}}\nonsensitive_m}_{\text{non-protected attributes}},  &\underbrace{\exists \sensitive_1,\dots, \exists \sensitive_n}_{\text{protected attributes}},
\phi_{\hat{Y}} \wedge (A=\mathbf{a}).\notag
\end{split}
\end{equation}
In $ \Phi_{\mathbf{a}} $, the existentially quantified variables $ \sensitive_1, \dots, \sensitive_n $ are assigned values  according  to the constraint $ A=\mathbf{a} $. \footnote{An RE-SSAT formula becomes an R-SSAT formula when the assignment to the existential variables are fixed.} Therefore, by solving the SSAT formula $ \Phi_{\mathbf{a}} $,  the SSAT solver finds the probability $ \Pr[\Phi_{\mathbf{a}}] $ for the protected group $ A=\mathbf{a} $ given the random values of $ \nonsensitive_1, \dots, \nonsensitive_m $, which is the PPV of the protected group $\mathbf{a} $ for the distribution $ \mathcal{D} $ and algorithm $\alg$. 

For simplicity, we have described computing the PPV of each compound protected group without considering the correlation between the protected and non-protected attributes. In reality, correlation exists between the protected and non-protected attributes. Thus, they may have different conditional distributions for different protected groups. We incorporate these conditional distributions in RE-SSAT encoding by evaluating the conditional probability $ p_i = \Pr[\nonsensitive_i =\text{TRUE}| A=\mathbf{a}] $ instead of the independent probability $\Pr[\nonsensitive_i =\text{TRUE}]$ for any $\nonsensitive_i\in \nonsensitive$. We illustrate this method in Example~\ref{example:re_ssat}.
%This relaxes the independence requirement.
\begin{example}[RE-SSAT encoding]
	\label{example:re_ssat}
	Here, we illustrate the RE-SSAT formula for calculating the PPV for the protected group `age $ \ge 40 $' in the decision tree of Figure~\ref{fig:fair_example}. We assign three Boolean variables $ F,I,J $ for the three nodes in the tree such that the literal $ F,I,J $ denote `fitness $ \ge 0.61 $', `income $ \ge 0.29 $', and `income $ \ge 0.69 $', respectively. We consider another Boolean variable $A$  where the literal $ A $ represents the protected group `age $ \ge 40 $'. Thus, the CNF formula  for the decision tree is $ (\neg F \vee I) \wedge (F \vee J) $. From the distribution in Figure~\ref{fig:fair_example}, we get $ \Pr[F] = 0.41, \Pr[I] = 0.93 $, and $ \Pr[J] = 0.09 $. Given this information, we calculate the PPV for the protected group `age $ \ge 40 $' by solving the RE-SSAT formula:
	\begin{equation}
	\Phi_A := \R^{0.41}F, \R^{0.93}I, \R^{0.09}J, \exists A, \; (\neg F \vee I) \wedge (F \vee J) \wedge A.\notag
	\end{equation}
	From the solution to this SSAT formula, we get $ \Pr[\Phi_A] = 0.43 $. Similarly, to calculate the PPV for the group `age $ < 40 $', we replace the unit (single-literal) clause $ A $ with $ \neg A $ in the CNF in $ \Phi_A $ and construct another SSAT formula $ \Phi_{\neg A} $ where $ \Pr[\Phi_{\neg A}] = 0.43 $. 
	Therefore, if $\Pr[F], \Pr[I], \Pr[J]$ are computed independently of $A$ and $\neg A$, both age groups demonstrate equal PPV as the protected attribute is not explicitly present in the classifier. 
	However, there is an implicit bias in the data distribution for different protected groups and the classifier unintentionally learns it. 
	To capture this implicit bias, we calculate the conditional probabilities  $ \Pr[F|A] = 0.01, \Pr[I|A] = 0.99 $, and $ \Pr[J|A] = 0.18 $ from the distribution. Using the conditional probabilities in  $\Phi_A $, we find that $ \Pr[\Phi_A] = 0.18 $ for `age $ \ge 40 $'. For `age $ < 40 $',  we similarly obtain $ \Pr[F|\neg A] = 0.82, \Pr[I|\neg A] = 0.88 $, and $ \Pr[J|\neg A] = 0.01 $, and thus  $ \Pr[\Phi_{\neg A}] = 0.72 $. 
	Thus, presented RE-SSAT encoding detects the discrimination of the classifier among different protected groups. An astute reader would observe that $I$ and $J$ are not independent. Following~\cite{chavira2008probabilistic}, we can simply capture relationship between the variables using constraints and if needed, auxiliary variables. In this case, it suffices to add the the constraint $J \rightarrow I$. 
\end{example}

\noindent\textbf{Measuring Fairness Metrics.}
As we compute the probability $\Pr[\hat{Y} = 1 | A = \mathbf{a}]$ by solving the SSAT formula $ \Phi_\mathbf{a} $, we  use $ \Pr[\Phi_\mathbf{a}] $ to measure different fairness metrics. 
For that, we compute $ \Pr[\Phi_\mathbf{a}] $ for all compound groups $\mathbf{a} \in A$ that requires solving exponential (with $n$) number of SSAT instances. 
We elaborate this enumeration approach, namely {\frameworkenum}, in Algorithm~\ref{algo:enum}  (Line~\ref{algo:justicia_enum_begin}--\ref{algo:justicia_enum_end}).

\begin{algorithm}[t!]
	\caption{\framework: SSAT-based Fairness Verifier}
	\label{algo:enum}
	\footnotesize
	\begin{algorithmic}[1]
		\Function{{\frameworkenum}}{$ X,A,\hat{Y} $}
		\label{algo:justicia_enum_begin}
		\State $ \phi_{\hat{Y}} := \mathsf{CNF}(\hat{Y} = 1) $
		%\State $ p_{i} = \mathsf{CalculateProb}(\nonsensitive_i), \forall \nonsensitive_i \in X $
		\ForAll{$\mathbf{a} \in A$ }
		\State $ p_{i} \leftarrow \mathsf{CalculateProb}(\nonsensitive_i | \mathbf{a}), \forall \nonsensitive_i \in X $
		\State $ \phi := \phi_{\hat{Y}} \wedge (A=\mathbf{a}) $
		\State $  \Phi_\mathbf{a} := \R^{p_{1}}\nonsensitive_1, \dots, \R^{p_{m}}\nonsensitive_m, \exists \sensitive_1,\dots, \exists \sensitive_n,  \phi $
		\State $ \Pr[\Phi_\mathbf{a}]  \leftarrow \mathsf{SSAT}(\Phi_\mathbf{a}) $ \Comment{returns a probability}
		\EndFor
		\State \Return $ \max_{\mathbf{a}} \; \Pr[\Phi_{\mathbf{a}}], \min_{\mathbf{a}} \; \Pr[\Phi_{\mathbf{a}}] $
		\label{algo:justicia_enum_end}
		\EndFunction

		\Function{{\frameworklearn}}{$ X,A,\hat{Y} $}
		\label{algo:justicia_learn_begin}
		\State $ \phi_{\hat{Y}} := \mathsf{CNF}(\hat{Y}  = 1) $
		\State $ p_{i} \leftarrow \mathsf{CalculateProb}(\nonsensitive_i), \forall \nonsensitive_i \in X $
		\State $  \Phi_\mathbf{ER} := \exists \sensitive_1,\dots, \exists \sensitive_n, \R^{p_{1}}\nonsensitive_1, \dots, \R^{p_{m}}\nonsensitive_m, \phi_{\hat{Y}} $
		\State $  \Phi'_\mathbf{ER} := \exists \sensitive_1,\dots, \exists \sensitive_n, \R^{p_{1}}\nonsensitive_1, \dots, \R^{p_{m}}\nonsensitive_m, \neg \phi_{\hat{Y}} $
		\State \Return $ \mathsf{SSAT}(\Phi_\mathbf{ER}), 1 - \mathsf{SSAT}(\Phi'_\mathbf{ER}) $
		\label{algo:justicia_learn_end}
		\EndFunction
	\end{algorithmic}

\end{algorithm}

We calculate the ratio of the minimum and the maximum probabilities according to the definition of disparate impact in Section~\ref{sec:preliminaries}. 
We compute statistical parity by taking the difference between the maximum and the minimum probabilities of all $ \Pr[\Phi_{\mathbf{a}}] $.
Moreover, to measure equalized odds, we compute two SSAT instances for each compound group with modified values of $ p_i $. 
Specifically, to compute TPR, we use the conditional probability $ p_i = \Pr[\nonsensitive_i|Y=1] $ on samples with class label $ Y = 1 $ and take the difference between the maximum and the minimum probabilities of all compound groups. In addition, to compute FPR, we use the conditional probability $ p_i = \Pr[\nonsensitive_i|Y=0] $ on samples with $ Y = 0 $ and take the difference similarly.
Thus, {\frameworkenum} allows us to compute different fairness metrics using a unified algorithmic framework.

\subsection{Learning Fairness with ER-SSAT Encoding}
\label{sec:learn_ssat}
In most practical problems, there can be exponentially many compound groups based on the different combinations of valuation to the protected attributes. 
Therefore, the enumeration approach in Section~\ref{sec:enumeration_ssat} may suffer from scalability issues. 
Hence, we propose efficient SSAT encodings to \textit{learn} the most favored group and the least favored group for given  $\alg$ and $ \mathcal{D} $, and to compute their PPVs to measure different fairness metrics. 

\textbf{Learning the Most Favored Group.}
In an SSAT formula $ \Phi $, the order of quantification of the Boolean variables in the prefix  carries distinct interpretation of the satisfying probability of $ \Phi $.  
In ER-SSAT formula, the probability of satisfying $ \Phi $ is the \textit{maximum} satisfying probability over the existentially quantified variables given the randomized quantified variables (by Rule 2, Sec.~\ref{sec:ssat}). 
In this paper, we leverage this property to compute the most favored group with the highest PPV. 
We consider the following ER-SSAT formula. 
\begin{equation}
\Phi_{\mathsf{ER}} := \exists \sensitive_1,\dots, \exists \sensitive_n,
 \R^{p_{1}}\nonsensitive_1, \dots, \R^{p_{m}}\nonsensitive_m,   \; \phi_{\hat{Y}}.
 \label{eq:er}
\end{equation}
%In the prefix of the ER-SSAT formula $ \Phi_{\mathsf{ER}} $, the protected variables are  existentially quantified and they are followed by randomized quantified non-protected variables, which is in reverse order  in Eq.~\eqref{eq:re}. 
The CNF formula $\phi_{\hat{Y}}$ is the CNF translation  of the classifier $ \hat{Y} = 1 $ without any specification of the compound protected group.  Therefore, as we solve $ \Phi_{\mathsf{ER}} $, we find the assignment to the existentially quantified variables $ A_1 = a^{\max}_1, \dots,A_n = a^{\max}_n $ for which the satisfying probability $ \Pr[\Phi_{\mathsf{ER}}] $ is maximum. 
Thus, we compute  the most favored group $ \mathbf{a}_{\mathsf{fav}} \triangleq \{ a^{\max}_1, \dots, a^{\max}_n \}$ achieving the highest PPV. 
%Since an assignment to $\{\sensitive_1, \dots, \sensitive_n\}$ uniquely denotes a compound protected group, say $ \mathbf{a}_{\mathsf{fav}} \in A $, we find the most favored group $ \mathbf{a}_{\mathsf{fav}} $ for which $ \Pr[\Phi_{\mathsf{ER}}] $ achieves the highest PPV. 

\textbf{Learning the Least Favored Group.}
In order to learn the least favored group in terms of PPV, we  compute the \textit{minimum} satisfying probability of the classifier $ \phi_{\hat{Y}} $ given the random values of the non-protected variables $ \nonsensitive_1, \dots, \nonsensitive_m $. In order to do so, we have to solve a `universal-random' (UR) SSAT formula (Eq.~\eqref{eq:ar}) with  universal quantification over the protected variables and randomized quantification over the non-protected variables (by Rule 3, Sec.~\ref{sec:ssat}).
\begin{equation}
\Phi_{\mathsf{UR}} := \forall \sensitive_1,\dots, \forall \sensitive_n,
\R^{p_{1}}\nonsensitive_1, \dots, \R^{p_{m}}\nonsensitive_m,   \; \phi_{\hat{Y}}.
\label{eq:ar}
\end{equation}
A UR-SSAT formula returns the minimum satisfying probability of $ \phi $ over the universally quantified variables in contrast to the ER-SSAT formula that returns the maximum satisfying  probability over the existentially quantified variables.  
Due to practical issues to solve UR-SSAT formula, in this paper, we leverage the \textit{duality} between UR-SSAT (Eq.~\eqref{eq:ar}) and ER-SSAT formulas (Eq.~\eqref{eq:er_complement}) 
\begin{equation}
\Phi'_{\mathsf{ER}} := \exists \sensitive_1,\dots, \exists \sensitive_n,
\R^{p_{1}}\nonsensitive_1, \dots, \R^{p_{m}}\nonsensitive_m,   \; \neg \phi_{\hat{Y}}.
\label{eq:er_complement}
\end{equation}
%\begin{equation}
%\Phi'_{\mathsf{ER}} = \underbrace{\exists a_1,\dots, \exists a_n}_{\text{protected attributes}},
%\underbrace{\R^{p_{1}}x_1, \dots, \R^{p_{m}}x_m}_{\text{non-protected attributes}},   \; \underbrace{\neg \phi_{\hat{Y}} }_{\phi}.
%\label{eq:er_complement}
%\end{equation}
and solve the UR-SSAT formula on the CNF $ \phi $ using the ER-SSAT formula on the complemented CNF $ \neg \phi $~\cite{littman2001stochastic}. Lemma~\ref{thm:dual} encodes this duality.
%We apply such a duality because of the practical unavailability of an UR-SSAT solver (to the best of our knowledge).  Formally, we solve the following ER-SSAT formula to find the least favored group. 
\begin{lemma}\label{thm:dual}
Given Eq.~\eqref{eq:ar} and~\eqref{eq:er_complement},	$ \Pr[\Phi_{\mathsf{UR}}] = 1 - \Pr[\Phi'_{\mathsf{ER}}]  $.
\end{lemma}
As we solve $\Phi'_{\mathsf{ER}}$, we obtain the assignment to the protected attributes $\mathbf{a}_{\mathsf{unfav}} \triangleq \{a^{min}_1, \dots, a^{min}_n\}$ that maximizes $\Phi'_{\mathsf{ER}}$. 
If $ p $ is the maximum satisfying probability of $ \Phi'_{\mathsf{ER}} $, according to Lemma~\ref{thm:dual}, $ 1 - p $ is the minimum satisfying probability of $ \Phi_{\mathsf{UR}} $,  which is the PPV of the least favored group $ \mathbf{a}_{\mathsf{unfav}}$. We present the algorithm for this learning approach, namely {\frameworklearn} in Algorithm~\ref{algo:enum} (Line~\ref{algo:justicia_learn_begin}--\ref{algo:justicia_learn_end}).

\iffalse
\begin{algorithm}[t!]
	\caption{\frameworklearn: Learning ER-SSAT Encoding}
	\label{algo:learn}
	\begin{algorithmic}[1]
		\Function{{\frameworklearn}}{$ X,A,\hat{Y} $}
		\State $ \phi_{\hat{Y}} = \mathsf{CNF}(\hat{Y}  = 1) $
		\State $ p_{i} = \mathsf{CalculateProb}(x_i), \forall x_i \in X $
		\State $  \Phi_\mathbf{ER} = \exists a_1,\dots, \exists a_n, \R^{p_{1}}x_1, \dots, \R^{p_{m}}x_m. \; \phi_{\hat{Y}} $
		\State $  \Phi'_\mathbf{ER} = \exists a_1,\dots, \exists a_n, \R^{p_{1}}x_1, \dots, \R^{p_{m}}x_m. \; \neg \phi_{\hat{Y}} $
		\State \Return $ \mathsf{SSAT}(\Phi_\mathbf{ER}), 1 - \mathsf{SSAT}(\Phi'_\mathbf{ER}) $
		\EndFunction
	\end{algorithmic}
\end{algorithm}
\fi

In ER-SSAT formula of Eq.~\eqref{eq:er_complement}, we need to negate the classifier $ \phi_{\hat{Y}} $ to another CNF formula $ \neg \phi_{\hat{Y}} $. The na\"ive approach of negating a CNF to another CNF generates exponential number of new clauses. Here, we can apply Tseitin transformation that increases the clauses linearly while introducing linear number of new variables~\cite{tseitin1983complexity}. As an alternative, we also directly encode the classifier $\alg$ for the negative class label $\hat{Y} = 0$ as a CNF formula and pass it to $\Phi'_{\mathsf{ER}} $, if possible. The last approach is generally more efficient than the other approaches as the resulting CNF is often smaller.

\begin{example}[ER-SSAT encoding]
	\label{example:er_ssat}
	Here, we illustrate the ER-SSAT encodings for learning the most favored and the least favored group in presence of multiple protected groups. As the example in Figure~\ref{fig:fair_example} is degenerate for this purpose, we introduce another protected group `sex $ \in $ \{male, female\}'. Consider a Boolean variable $ S $ for `sex' where the literal $ S $ denotes `sex = male'. With this new protected attribute, let the classifier be  $\alg \triangleq (\neg H \vee I \vee S) \wedge (H \vee J)$, where $ A,H,I,J $ have same distributions as discussed in Example~\ref{example:re_ssat}. 
	Hence, we obtain the ER-SSAT formula of $\alg$ to learn the most favored group:
	\begin{equation}
	\Phi_{\mathsf{ER}} =  \exists S,\exists A, \R^{0.41}H, \R^{0.93}I, \R^{0.09}J, \; (\neg H \vee I \vee S) \wedge (H \vee J).\notag
	\end{equation}
	As we solve $ \Phi_{\mathsf{ER}} $, we learn that the assignment to the existential variables $ \sigma(S) = 1, \sigma(A) = 0$, i.e. `male individuals with age $ < 40 $' is the most favored group with PPV computed as $ \Pr[\Phi_{\mathsf{ER}}] = 0.46$. Similarly, to learn the least favored group, we negate the CNF of the classifier $\alg$ to obtain the following ER-SSAT formula:
	\begin{equation}
	\Phi_{\mathsf{ER'}} =  \exists S, \exists A, \R^{0.41}H, \R^{0.93}I, \R^{0.09}J, \; \neg((\neg H \vee I \vee S) \wedge (H \vee J)).\notag
	\end{equation}
	Solving $ \Phi_{\mathsf{ER'}} $, we learn the assignment $ \sigma(S) = 0, \sigma(A) = 0  $ and $  \Pr[\Phi_{\mathsf{ER'}}] = 0.57 $. Thus, `female individuals with age $ < 40 $' constitute the least favored group with PPV:  $ 1-0.57 = 0.43$. 
	Thus, {\frameworklearn} allows us to learn the most and least favored groups and the corresponding discrimination.
\end{example}
We use the PPVs of the most and least favored groups to compute fairness metrics as described in Section~\ref{sec:enumeration_ssat}. 
%The following lemma states the duality between ER-SSAT and UR-SSAT formulas.
%\begin{lemma}
%	\blue{Solving the RE-SSAT problem is} $\mathrm{NP}^{\mathrm{PP}}$ hard.
%\end{lemma}
We prove equivalence of {\frameworkenum} and {\frameworklearn} in Lemma~\ref{lm:equivalence}.
\begin{lemma}
	\label{lm:equivalence}
	Let $ \Phi_{\mathbf{a}} $ be the RE-SSAT formula for computing the PPV of the compound protected group $ \mathbf{a} \in A $. If $ \Phi_{\mathsf{ER}} $ is the ER-SSAT formula for learning the most favored group and $ \Phi_{\mathsf{UR}} $ is the UR-SSAT formula for learning the least favored group, then
	$\max_{\mathbf{a}} \; \Pr[\Phi_{\mathbf{a}}] = \Pr[\Phi_{\mathsf{ER}}]$   
	and
	$\min_{\mathbf{a}} \; \Pr[\Phi_{\mathbf{a}}] = \Pr[\Phi_{\mathsf{UR}}]$.   
\end{lemma}

\subsection{Theoretical Analysis: Error Bounds}\label{sec:theory}
We access the data generating distribution through  finite number of samples observed from it. These finite sample set introduce errors in the computed probabilities of the randomised quantifiers being $1$. These finite-sample errors in computed probabilities induce further errors in the computed positive predictive value (PPV) and fairness metrics. In this section, we provide a bound on this finite-sample error.

Let us consider that $\hat{p_i}$ is the estimated probability of a Boolean variable $ \bool_i $ being assigned to $ 1 $ from $k$-samples and $p_i$ is the true probability according to $ \mathcal{D} $. 
%If $ | p_i - \hat{p_i}| \le \epsilon $, i.e., $ \epsilon $ additive error for small $ \epsilon \approx 0 $, we want to compute the error of probability $ p $ of the satisfaction of the SSAT formula $ \Phi $.  
Thus, the true satisfying probability $p$ of $ \Phi $ is the weighted sum of all satisfying assignments of the CNF $ \phi $: $p = \sum_\sigma \prod_{\bool_i \in \sigma}p_i$.
This probability is estimated as $\hat{p}$ using $k$-samples from the data generating distribution $\mathcal{D}$ such that $\hat{p} \leq \epsilon_0 p$ for $\epsilon_0 \geq 1$. 
%However, the true satisfying probability $ \hat{p} $  of $ \Phi $ is 
%\[
%\hat{p} = \sum_\sigma \prod_{\bool_i \in \sigma}\hat{p_i}  \leq \sum_\sigma \prod_{\bool_i \in \sigma}(1 \pm \epsilon_i) p_i = \prod_{i =1}^m (1 \pm \epsilon_i) p = \epsilon_0 p
%\]
\iffalse
If we consider $\epsilon_i = \frac{\ln \epsilon_0}{2^i}$, we obtain
\begin{equation*}
\begin{split}
\prod_{i =1}^m (1 \pm \epsilon_i) &\leq \left(\frac{1}{n} \sum_{i} (1 + \epsilon_i)\right)^n \\
&= (\frac{1}{n} \sum_{i} (1+\frac{\ln \epsilon_0}{2^i})^n\\
&\leq (1  +\frac{\ln \epsilon_0}{n})^n\\
&\leq e^{\ln \epsilon_0} = \epsilon_0.\notag
\end{split}
\end{equation*}
\fi
\begin{theorem}\label{thm:sample}
	For an ER-SSAT problem, the sample complexity is given by 
	$ k = O\left((n+ \ln(1/\delta))\frac{\ln m}{\ln \epsilon_0} \right)$,
	where $\frac{\hat{p}}{p} \leq \epsilon_0$ with probability $1-\delta$ such that $\epsilon_0 \geq 1$.
\end{theorem}
%Theorem~\ref{thm:sample} states that for the prescribed number of samples the estimated disparate impact $\hat{DI}$ and statistical parity $\hat{SP}$ would also satisfy $\hat{DI} \leq  \epsilon_0 DI, $ and $\hat{SP} \leq 2\epsilon_0 SP$.
%Apply Hoeffding inequality for each term and then use Union bound for $2^n$ existential variables.
\begin{corollary}
	\label{cor:error}
	If $k$ samples are considered from the data-generating distribution in {\framework} such that 
	$
	k = O\left((n+ \ln(1/\delta))\frac{\ln m}{\ln \epsilon_0}\right),
	$
	the estimated disparate impact $\hat{DI}$ and statistical parity $\hat{SP}$ satisfy, with probability $1-\delta$,
	$
	\hat{DI} \leq  \epsilon_0 DI, \quad \text{and} \quad \hat{SP} \leq 2\epsilon_0 SP.
	$
\end{corollary}
This implies that given a classifier $ \alg \triangleq \Pr(\hat{Y}|X, A) $ represented as a CNF formula and a data-generating distribution $ (X,A,Y) \sim \mathcal{D} $, {\framework} can verify independence and separation notion of fairness up to an error level $\epsilon_0$ and $2\epsilon_0$ with probability $1-\delta$.
Thus, {\framework} is a sound framework of fairness verification with high probability.
%\red{Proof of all theorems for arxiv version}
%Use definition of $GF = p_1/p_2$.

\iffalse

\begin{corollary}[Hypothesis]
	If $k$ samples are considered in \framework and the estimated group fairness value $\hat{GF}$ satisfies
	\begin{equation*}
	1- e^{\epsilon_0} \leq \frac{\hat{GF}}{GF} \leq 1 + e^{\epsilon_0}
	\end{equation*}
	with probability $1-\delta$, then
	\begin{equation*}
	k = O\left((n+ \ln(1/\delta))\frac{m + \ln m}{\ln \epsilon_0}\right).
	\end{equation*}
\end{corollary}
\fi

\subsection{Practical Settings}
\label{sec:practical-setting}
In this section, we relax the assumptions on access to Boolean classifiers and Boolean attributes, and extend {\framework} to verify fairness metrics for more practical settings of decision trees, linear classifiers, and continuous attributes.

\paragraph{Extending to Decision Trees and Linear Classifiers. }
In the SSAT approach of Section~\ref{sec:framework}, we assume that the classifier $\alg$ is represented as a CNF formula.  
We extend {\framework} beyond CNF classifiers to decision trees and linear classifiers, which are widely used in the fairness studies~\cite{zemel2013learning,raff2018fair,zhang2019faht}.
%{\color{red}In the literature of interpretable machine learning, several studies have been conducted for learning CNF classifiers in the supervised learning setting, which include but are not limited to the work of~\cite{angelino2017learning,malioutov2018mlic,ghosh19incremental}.  We leverage these techniques. (not needed)}

\textit{Binary decision trees} are trivially encoded as  CNF formulas.  In the binary decision tree, each node in the tree is a literal. A \textit{path from the root to the leaf} is a conjunction of literals and thus, a \textit{clause}. The \textit{tree} itself is a disjunction of all paths and thus, a \textit{DNF (Disjunctive Normal Form)}. In order to derive a CNF of a decision tree, we first construct a DNF by including all paths terminating at leaves with negative class label ($ \hat{Y} = 0 $) and then complement the DNF to CNF using De Morgan's rule. 

\textit{Linear classifiers on Boolean attributes} are encoded into CNF formulas using pseudo-Boolean encoding~\cite{philipp2015pblib}. We consider a linear classifier  $ W^T X + b \ge 0 $ on Boolean attributes $ X $ with weights $ W \in \mathbb{R}^{|X|} $ and bias $ b \in \mathbb{R} $.  We first normalize $W$ and $b $ in $ [-1,1] $ and then round to integers so that the decision boundary becomes a pseudo-Boolean constraint.  We then apply  pseudo-Boolean constraints to CNF translation to encode the decision boundary to CNF. This encoding usually introduces additional Boolean variables and results in large CNF. In order to generate a smaller CNF, we can trivially apply thresholding  on the weights to consider attributes with higher weights only. For instance, if the weight $  |w_i| \le \lambda $ for a threshold $ \lambda  \in \mathbb{R}^+$ and $ w_i \in W $, we can set $ w_i = 0 $. Thus, the attributes with lower weights and thus, less importance do not appear in the encoded CNF.  Moreover, all introduced variables in this CNF translation are given existential ($ \exists $) quantification and they appear in the inner-most position in the prefix of the SSAT formula. Thus, the presented ER-SSAT formulas become effectively ERE-SSAT formulas.

\paragraph{Extending to Continuous Attributes.}
In practical problems, attributes are generally real-valued or categorical but classifiers, which are naturally expressed as CNF such as~\cite{GMM20}, are generally trained on a Boolean abstraction of the input attributes.
In order to perform this Boolean abstraction, each categorical attribute is one-hot encoded and each real-valued attribute is discretised into a set of Boolean attributes~\cite{LKCL2019,GMM20}. 

For a binary decision tree, each attribute, including the continuous ones, is compared against a constant at each internal node of the tree. We fix a Boolean variable for each internal node, where the Boolean assignment to the variable decides one of the two branches to choose from the current node.  

Linear classifiers are generally trained on continuous attributes, where we apply the following discretization. 
Let us consider a continuous attribute $X_c$, where $w$ is its weight during training. 
We discretize $ X_c $ to a set $ \mathbf{B} $ of Boolean attributes and recalculate the weight of each variable in $ \mathbf{B} $ based on $ w $. 
For the discretization of $X_c$, we consider the interval-based approach\footnote{Our implementation is agnostic to any discretization technique.}. 
For each interval in the continuous space of $X_c$, we consider a Boolean variable $B_i \in \mathbf{B}$, such that $ B_i $ is assigned TRUE when the attribute-value of $X_c$ lies within the $i^{\mathrm{th}}$ interval and $ B_i $ is assigned FALSE otherwise. 
Following that, we assign the weight of $ B_i $ to be $ \mu_i\times w $, when $ \mu_i $ is the mean of the $i^{\mathrm{th}}$ interval and  $ B_i $ is TRUE. 
We can show that if we consider infinite number of intervals, $ X_c \approx \sum_i \mu_i B_i $.

\section{Empirical Performance Analysis}
\label{sec:experiments}
In this section, we discuss the empirical studies to evaluate the performance of {\framework} in verifying different fairness metrics. We first discuss the experimental setup and the objective of the experiments and then evaluate the experimental results.
\subsection{Experimental Setup}
We have implemented a prototype of {\framework} in Python (version $ 3.7.3 $). The core computation of {\framework} relies on solving SSAT formulas using an off-the-shelf SSAT solver. To this end, we employ the state of the art RE-SSAT solver of~\cite{lee2017solving} and the ER-SSAT solver of~\cite{lee2018solving}. Both solvers output the exact  satisfying probability of the SSAT formula. 

For comparative evaluation of {\framework}, we have experimented with two state-of-the-art distributional verifiers FairSquare and VeriFair, and also a sample-based fairness measuring tool: AIF360. 
In the experiments, we have studied three type of classifiers: CNF learner, decision trees and logistic regression classifier.
Decision tree and logistic regression are implemented using scikit-learn module of Python~\cite{PVGMTGB2011} and we use the MaxSAT-based CNF learner IMLI of~\cite{ghosh19incremental}. We have used the PySAT library~\cite{imms-sat18} for encoding the decision function of the logistic regression classifier into a CNF formula.
We have also verified two fairness-enhancing algorithms: reweighing algorithm~\cite{kamiran2012data} and the optimized pre-processing  algorithm~\cite{calmon2017optimized}. 
We have experimented on multiple datasets containing multiple protected attributes: the UCI Adult and German-credit dataset~\cite{DK2017uci},  ProPublica’s COMPAS recidivism dataset~\cite{angwin2016machine}, Ricci dataset~\cite{mcginley2010ricci}, and Titanic dataset\footnote{\url{https://www.kaggle.com/c/titanic}}.
\iffalse 
Since both {\framework}  and FairSquare take a  probability distribution of the attributes as input, we perform five-fold cross validation, use the train set for learning the classifier, compute distribution on the test set and finally verify fairness metrics such as disparate impact and statistical parity difference on the distribution. 
\fi
%that pre-process the dataset to mitigate its bias

Our empirical studies have the following objectives:

\begin{enumerate}
	\item How accurate and scalable {\framework} is with respect to existing fairness verifiers, FairSquare and VeriFair?
	\item Can {\framework} verify the effectiveness of different fairness-enhancing algorithms on different datasets?
	\item Can {\framework} verify fairness in the presence of compound sensitive groups?
	\item How robust is {\framework} in comparison to sample-based tools like AIF360 for varying sample sizes?
	\item How do the computational efficiencies of {\frameworklearn} and {\frameworkenum} compare?
\end{enumerate}

%\begin{enumerate}
%	\item How accurate and scalable {\framework} is with respect to existing fairness verifiers, FairSquare and VeriFair?
%	\item Can {\framework} verify the effectiveness of different fairness-enhancing algorithms on different datasets?
%	\item Can {\framework} verify fairness in the presence of compound sensitive groups?
%	\item How robust is {\framework} in comparison to empirical tools like AIF360 for varying sample sizes?
%\end{enumerate}

Our experimental studies validate that {\framework} is more accurate and scalable than the state-of-the-art verifiers FairSquare and VeriFair. {\framework} is able to verify the effectiveness of different fairness-enhancing algorithms for multiple fairness metrics, and datasets. {\framework} achieves scalable performance in the presence of compound sensitive groups that the existing verifiers cannot handle.  {\framework} is also more robust than the sample-based tools such as AIF360.
Finally, {\frameworklearn} is significantly efficient in terms of runtime than {\frameworkenum}.
\begin{table}[t!]
    \centering
        \caption{Results on synthetic benchmark.  `\textemdash'~ refers that the verifier cannot compute the metric. }
        \label{tab:synthetic}
        \vspace*{-.2em}
        \setlength{\tabcolsep}{.1em}
            \begin{tabular}{ccccccc}
                \toprule
                Metric & Exact  & {\framework} & FairSquare & VeriFair & AIF360\\
                 \midrule
				Disparate impact &  $ 0.26 $  &  $ 0.25 $    &  $ 0.99 $  &  $ 0.99 $  &  $ 0.25 $  \\
				Stat. parity &  $ 0.53 $  &  $ 0.54 $    & \textemdash & \textemdash &  $ 0.54 $  \\
                \bottomrule
    \end{tabular}
\end{table}

\subsection{Experimental Analysis}

\begin{comment}
\begin{figure}[t]
\centering
\includegraphics[scale=.32]{figures/sanity_DI_DT.pdf}
\caption{Results of different verifiers on  a synthetic problem.
}
\label{fig:synthetic}
\end{figure}
\end{comment}

%\subsubsection{Performance of Different Verifiers.}
\paragraph{Accuracy: Less Than $  1\%$-error.} 
In order to assess the accuracy of different verifiers, we have considered the decision tree in Figure~\ref{fig:fair_example} for which the fairness metrics  are analytically computable. 
In Table~\ref{tab:synthetic}, we show the computed fairness metrics by {\framework}, FairSquare, VeriFair, and AIF360. We observe that {\framework} and AIF360  yield more accurate estimates of DI and SP compared against the ground truth with less than $1\%$ error.
FairSquare and VeriFair  estimate the disparate impact to be $0.99$ and thus, being unable to verify the fairness violation. 
Thus, {\framework} is significantly accurate than the existing formal verifiers: FairSquare and VeriFair. 
%First we observe that both RE and ER encoding result in the same disparate impact, thereby showing the equivalence between the two encodings.  

\begin{table}[t!]
    \centering
        \caption{Scalability of different verifiers in terms of execution time (in seconds).  DT and LR refer to decision tree and logistic regression respectively. `\textemdash'~ refers to timeout.}
        \label{tab:FS_VF_Justicia}
        \setlength{\tabcolsep}{.3em}
        \vspace*{-.3em}
            \begin{tabular}{lrrrrrrrr}
                \toprule
                Dataset   & \multicolumn{2}{c}{Ricci} & \multicolumn{2}{c}{Titanic} & \multicolumn{2}{c}{COMPAS} &  \multicolumn{2}{c}{Adult} \\ 
                \cmidrule(lr){2-3}
                \cmidrule(lr){4-5}
                \cmidrule(lr){6-7}
                \cmidrule(lr){8-9}

                Classifier & DT     & LR  & DT & LR  & DT  & LR  & DT  & LR \\ \midrule

{\framework} &  $ 0.1  $  &  $ 0.2  $  &  $ 0.1  $  &  $ 0.9  $  &  $ 0.1  $  &  $ 0.2  $  &  $ 0.2  $  &  $ 1.0  $  \\
FairSquare &  $ 4.8  $  & \textemdash &  $ 16.0  $  & \textemdash &  $ 36.9  $  & \textemdash & \textemdash & \textemdash \\
VeriFair &  $ 5.3  $  &  $ 2.2  $  &  $ 1.2  $  &  $ 0.8  $  &  $ 15.9  $  &  $ 11.3  $  &  $ 295.6  $  &  $ 61.1  $  \\
                \bottomrule
    \end{tabular}\vspace*{-1em}
\end{table}

\paragraph{Scalability: $ 1 $ to $ 3 $ Orders of Magnitude Speed-up.} 
%Since {\framework} appears to be more accurate than FairSquare and VeriFair in the synthetic benchmark, 
We have tested the scalability of {\framework}, FairSquare, and VeriFair on practical benchmarks with a timeout of $900$ seconds and reported the execution time of these verifiers on decision tree and logistic regression in Table~\ref{tab:FS_VF_Justicia}. We observe that {\framework} shows impressive scalability than the competing verifiers. Particularly, {\framework} is $ 1 $ to $ 2 $ orders of magnitude faster than FairSquare and  $ 1 $ to $ 3 $ orders of magnitude faster than VeriFair. Additionally, FairSquare times out in most  benchmarks.
Thus, {\framework} is not only accurate but also scalable than the existing verifiers.

\begin{figure}[t!]
	%		%%%\vspace*{-.5em}
	\centering
	\begin{minipage}{0.33\columnwidth}
		\includegraphics[scale=.2]{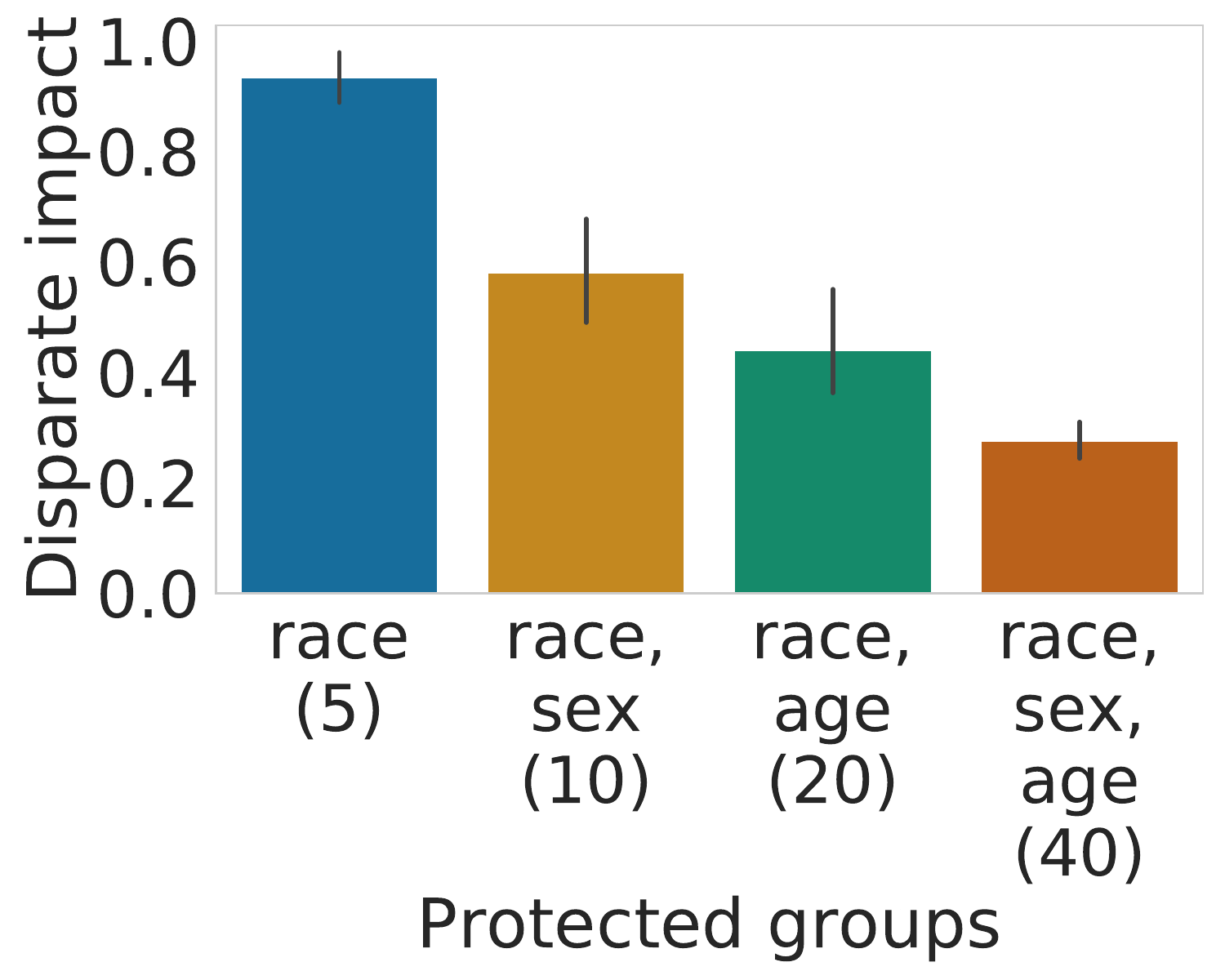}
	\end{minipage}\hfill
	\begin{minipage}{0.33\columnwidth}
		\includegraphics[scale=.2]{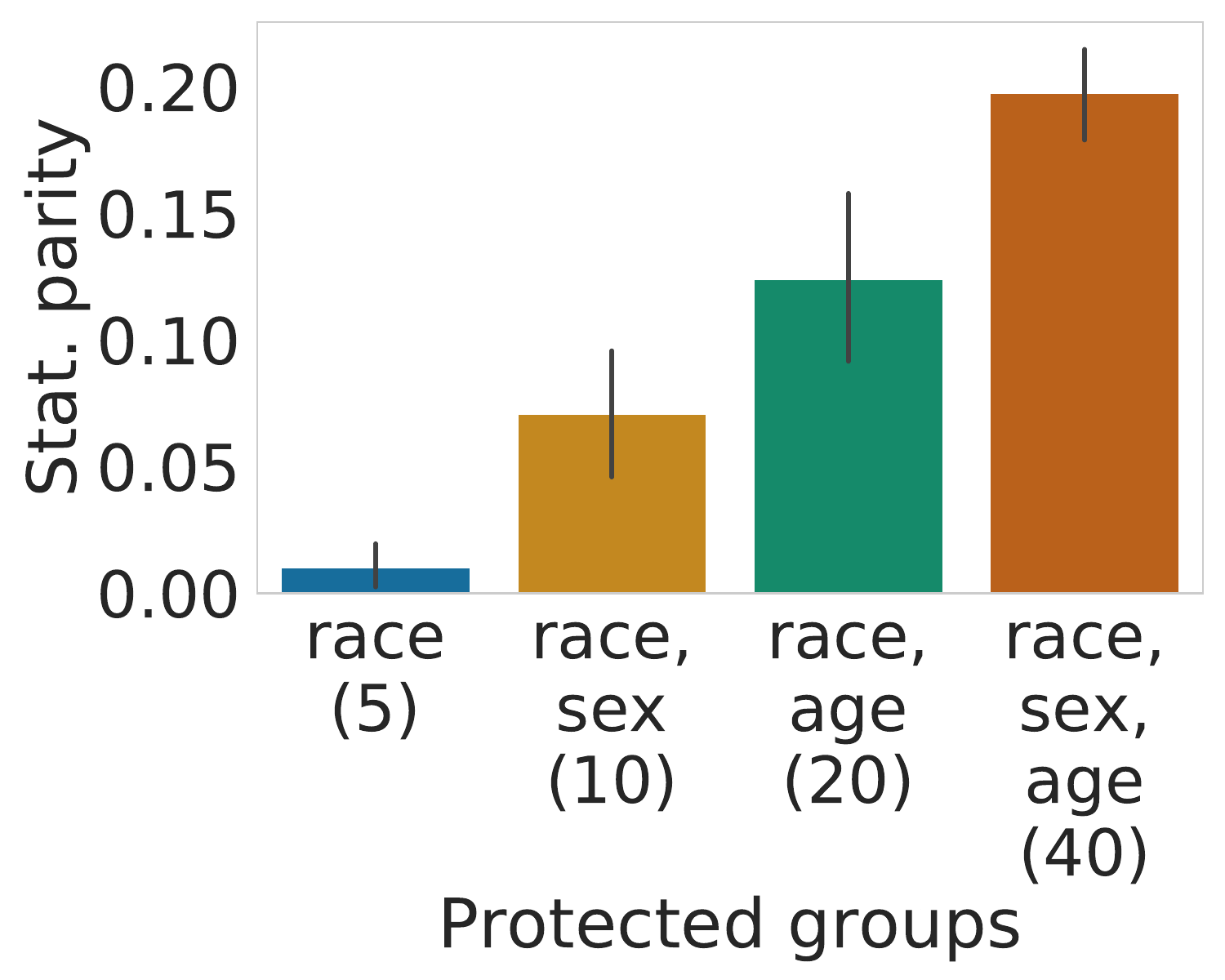}
	\end{minipage}\hfill
	\begin{minipage}{0.33\columnwidth}
		\includegraphics[scale=.2]{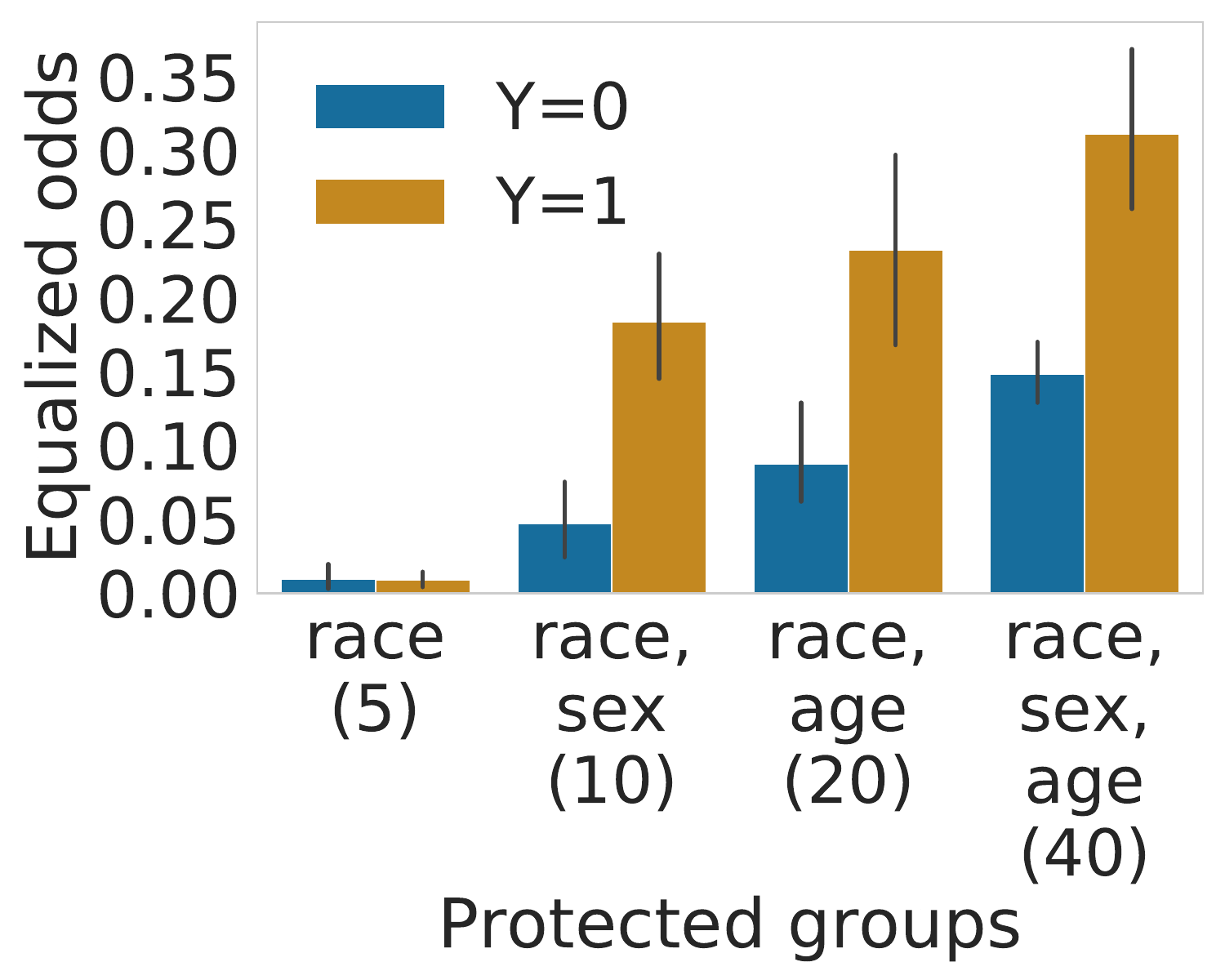}
	\end{minipage}
	%		%%%\vspace*{-.5em}
	\caption{Fairness metrics measured by {\framework} for different protected groups in the Adult dataset. The number within parenthesis in the xticks denotes total compound groups.}
	\label{fig:protected_groups}
	
\end{figure}

\paragraph{Verification: Detecting Compounded Discrimination in Protected Groups.}
We have tested {\framework} for datasets consisting of multiple protected attributes and reported the results in Figure~\ref{fig:protected_groups}. {\framework} operates on datasets with even 40 compound protected groups and can potentially scale more than that while the state-of-the-art fairness verifiers (e.g., FairSquare and VeriFair) consider a single protected attribute.
Thus, {\framework} removes an important limitation in practical fairness verification. 
Additionally, we observe in most datasets the disparate impact decreases and thus, discrimination increases as more compound protected groups are considered. For instance, when we increase the total  groups from $ 5 $ to $ 40 $ in the Adult dataset, disparate impact decreases from around $ 0.9 $ to $ 0.3 $, thereby detecting higher discrimination. Thus, {\framework} detects that the marginalized individuals of a specific type (e.g., `race')  are even more discriminated and marginalized when they also belong to a marginalized group of another type (e.g., `sex').

\newcommand{\STAB}[1]{\begin{tabular}{@{}c@{}}#1\end{tabular}}
\begin{table*}       
    \centering
        \caption{Verification of different fairness enhancing algorithms for multiple datasets and classifiers using {\framework}. Numbers in bold refer to fairness improvement  compared against the unprocessed (orig.) dataset. RW and OP refer to reweighing and optimized-preprocessing algorithm respectively. Results for German-credit dataset is in Appendix~\ref{sec:exp_extra} }\label{tab:fair_algo_verification}
        \setlength{\tabcolsep}{.2em}
            \begin{tabular}{ll
            				ccc
            				ccc
%            				ccc
%            				ccc
            				ccc
            				ccc}
                \toprule
                \multirow{3}{*}{Classifier}& Dataset $ \rightarrow $   & 
                \multicolumn{6}{c}{Adult} &
%                \multicolumn{6}{c}{German} &
                \multicolumn{6}{c}{COMPAS} \\ 
                \cmidrule(lr){3-8}
                \cmidrule(lr){9-14}
%                \cmidrule(lr){15-20}
                & Protected  $ \rightarrow $ & 
                \multicolumn{3}{c}{Race}   & \multicolumn{3}{c}{Sex}  &
%                \multicolumn{3}{c}{Age}   & \multicolumn{3}{c}{Sex}  &
                \multicolumn{3}{c}{Race}   & \multicolumn{3}{c}{Sex}
                \\ 
                \cmidrule(lr){3-5}
                \cmidrule(lr){6-8}
                \cmidrule(lr){9-11}
                \cmidrule(lr){12-14}
%                \cmidrule(lr){15-17}
%                \cmidrule(lr){18-20}

                 & Algorithm  $ \rightarrow $ &  
                orig. & RW & OP & 
                orig. & RW & OP &
%                orig. & RW & OP &
%                orig. & RW & OP &
                orig. & RW & OP &
                orig. & RW & OP \\ 
                \midrule

              \multirow{3}{*}{\shortstack{Logistic \\ regression}}
              & Disparte impact&  $ 0.23 $ &  $ \mathbf{0.85} $ &  $ \mathbf{0.59} $ &  $ 0.03 $ &  $ \mathbf{0.61} $ &  $ \mathbf{0.62} $ &  $ 0.34 $ &  $ \mathbf{0.36} $ &  $ \mathbf{0.47} $ &  $ 0.48 $ &  $ \mathbf{0.80} $ &  $ \mathbf{0.74} $  \\
              & Stat. parity&  $ 0.09 $ &  $ \mathbf{0.01} $ &  $ \mathbf{0.05} $ &  $ 0.16 $ &  $ \mathbf{0.04} $ &  $ \mathbf{0.03} $ &  $ 0.39 $ &  $ \mathbf{0.33} $ &  $ \mathbf{0.21} $ &  $ 0.23 $ &  $ \mathbf{0.09} $ &  $ \mathbf{0.10} $  \\
              & Equalized odds&  $ 0.13 $ &  $ \mathbf{0.03} $ &  $ \mathbf{0.10} $ &  $ 0.30 $ &  $ \mathbf{0.02} $ &  $ \mathbf{0.06} $ &  $ 0.38 $ &  $ \mathbf{0.33} $ &  $ \mathbf{0.18} $ &  $ 0.17 $ &  $ 0.19 $ &  $ \mathbf{0.07} $  \\
              \midrule
              \multirow{3}{*}{\shortstack{Decision \\ tree}}
              & Disparte impact&  $ 0.82 $ &  $ 0.60 $ &  $ 0.67 $ &  $ 0.00 $ &  $ \mathbf{0.73} $ &  $ \mathbf{0.95} $ &  $ 0.61 $ &  $ 0.58 $ &  $ 0.57 $ &  $ 0.94 $ &  $ 0.78 $ &  $ 0.63 $  \\
              & Stat. parity&  $ 0.02 $ &  $ 0.05 $ &  $ 0.04 $ &  $ 0.14 $ &  $ \mathbf{0.05} $ &  $ \mathbf{0.01} $ &  $ 0.18 $ &  $ \mathbf{0.17} $ &  $ \mathbf{0.17} $ &  $ 0.02 $ &  $ 0.09 $ &  $ 0.18 $  \\
              & Equalized odds&  $ 0.07 $ &  $ \mathbf{0.05} $ &  $ \mathbf{0.03} $ &  $ 0.47 $ &  $ \mathbf{0.03} $ &  $ \mathbf{0.04} $ &  $ 0.17 $ &  $ \mathbf{0.16} $ &  $ \mathbf{0.16} $ &  $ 0.07 $ &  $ \mathbf{0.05} $ &  $ 0.16 $  \\

                \bottomrule
    \end{tabular}
\end{table*}

\paragraph{Verification: Fairness of Algorithms on Datasets.}
We have experimented with two fairness-enhancing algorithms: the reweighing (RW) algorithm and the optimized-preprocessing (OP) algorithm.
Both of them pre-process to remove statistical bias from the dataset. 
We study the effectiveness of these algorithms using {\framework} on three datasets each with two different protected attributes.  
In Table~\ref{tab:fair_algo_verification}, we report different fairness metrics on logistic regression and decision tree. We observe that {\framework} verifies fairness improvement as the bias mitigating algorithms are applied.  For example, for the Adult dataset with `race' as the protected attribute, disparate impact increases from $ 0.23 $ to $ 0.85 $ for applying the reweighing algorithm on logistic regression classifier. In addition, statistical parity decreases from $ 0.09 $ to $ 0.01 $, and equalized odds decreases from $ 0.13 $ to $ 0.03 $, thereby showing the effectiveness of reweighing algorithm in all three fairness metrics. 
{\framework} also finds instances where the fairness algorithms fail, specially when considering the decision tree classifier. 
Thus, {\framework} enables verification of different fairness enhancing algorithms in literature.

\begin{figure}[t!]
	\centering
	\includegraphics[scale=.2]{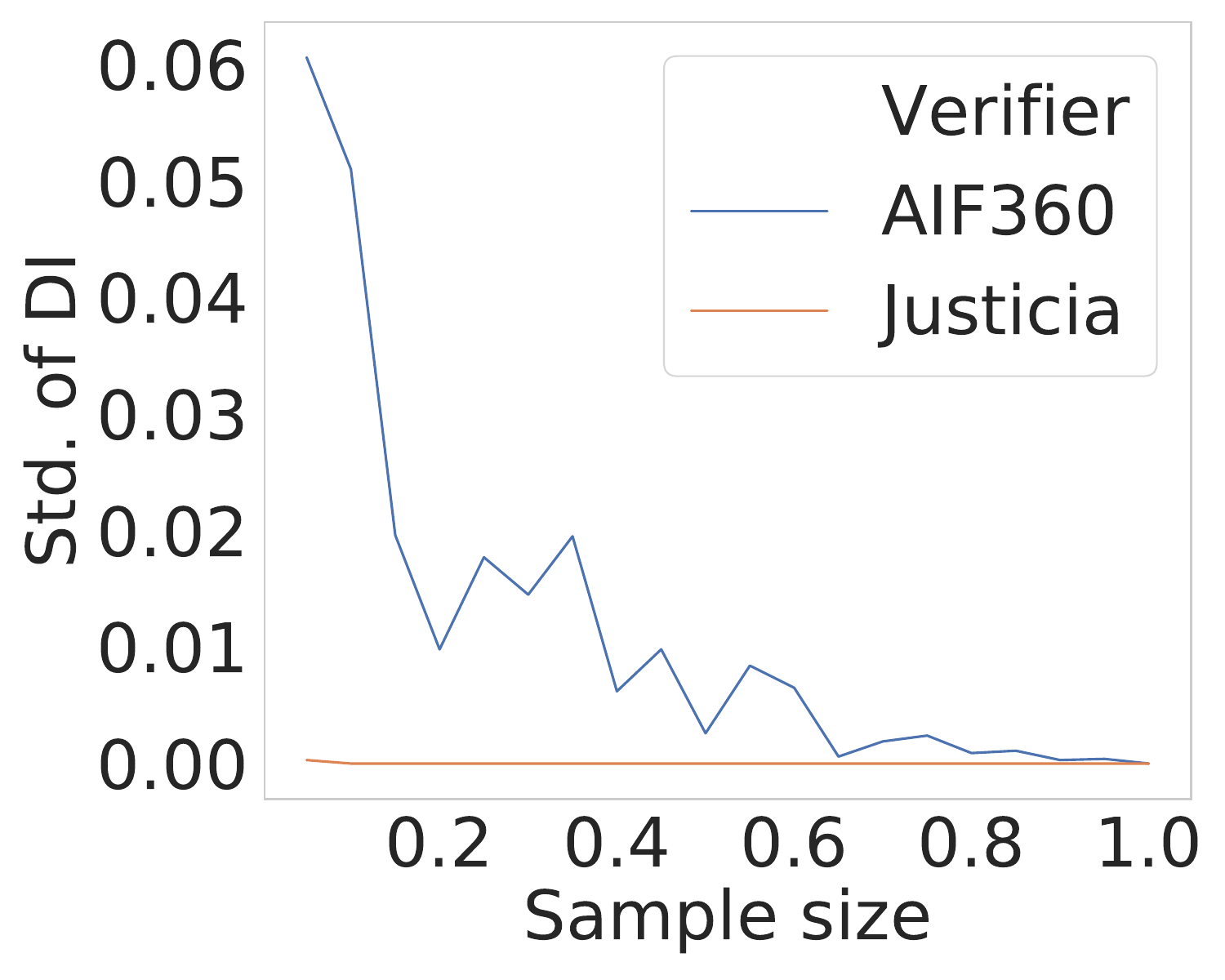}
	\includegraphics[scale=.2]{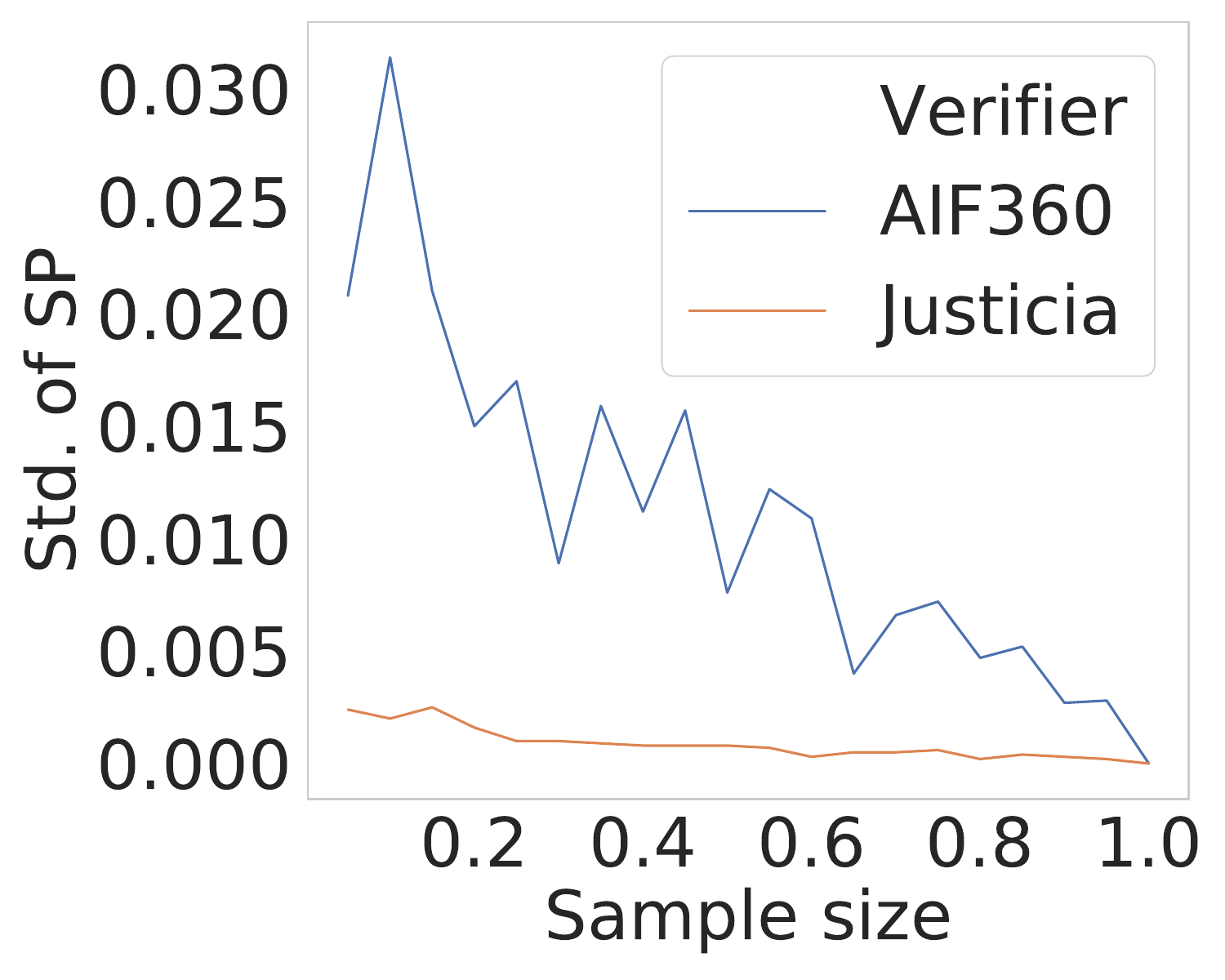}
	%		%%%\vspace*{-.5em}
	\caption{Standard deviation in estimation of disparate impact (DI) and stat. parity (SP)  for different sample sizes. {\framework} is more robust with variation of sample size than  AIF360. }
	\label{fig:sample-size}
	
\end{figure}

\paragraph{Robustness: Stability to Sample Size.} 
We have compared the robustness of {\framework} with AIF360 by varying the sample-size and reporting the standard deviation of different fairness metrics. 
In Figure~\ref{fig:sample-size}, AIF360 shows higher standard deviation for lower sample-size and the value decreases as  the sample-size increases. 
In contrast, {\framework} shows significantly lower ($\sim10\times$ to $100\times$) standard deviation for different sample-sizes. 
The reason is that AIF360 empirically measures on a fixed test dataset whereas {\framework} provides estimates over the data generating distribution.
Thus, {\framework} is more robust than the sample-based verifier AIF360.

\begin{figure}[t!]
	\begin{center}
		\includegraphics[scale=.4]{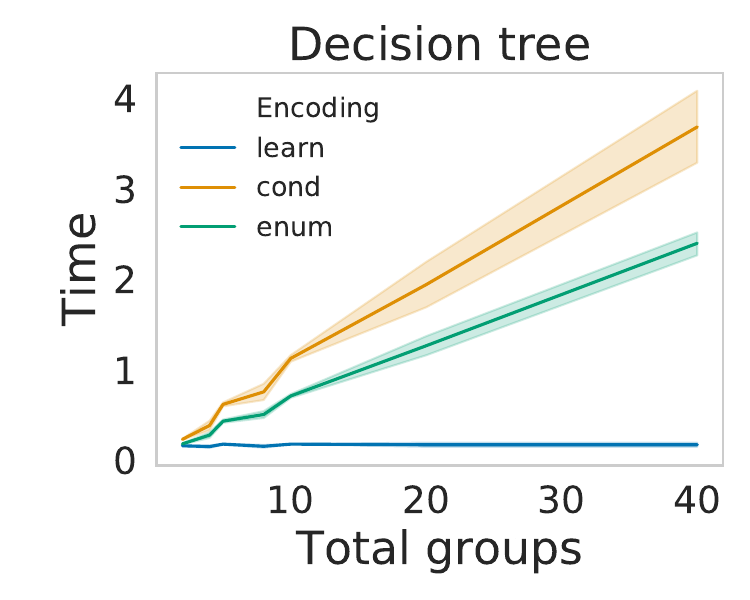}
		\includegraphics[scale=.4]{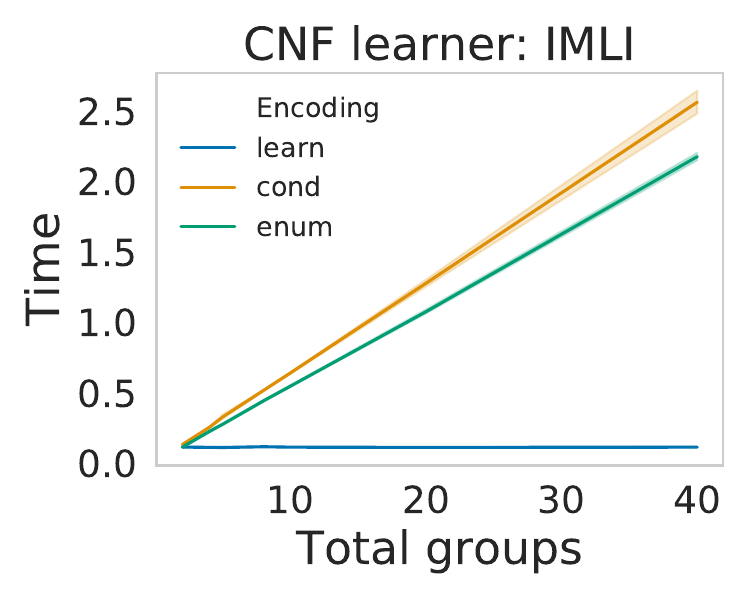}
		\hfill
		%%%\vspace*{-1em}
		\caption{Runtime comparison of different encodings while varying total protected groups in the Adult dataset.}
		\label{fig:runtime_diff_encodings}
	\end{center}
	%%%\vspace*{-1.5em}
\end{figure}

\paragraph{Comparative Evaluation of Different Encodings.}
While both {\frameworkenum} and {\frameworklearn}  have the same output according to Lemma~\ref{lm:equivalence},  {\frameworklearn} encoding  improves exponentially  in runtime  than {\frameworkenum} encoding on both decision tree and Boolean CNF classifiers as we vary the total compound groups in Figure~\ref{fig:runtime_diff_encodings}. {\frameworkcond} also has an exponential trend in runtime similar to {\frameworkenum}.  This analysis justifies that the na\"ive enumeration-based approach cannot verify large-scale fairness problems containing multiple protected attributes, and {\frameworklearn} is a more efficient approach for practical use.

%The runtime efficiency of {\frameworklearn} posits it as a more scalable and practical approach to verify fairness than {\frameworkenum}.
%\red{Justify the use of }.

\section{Discussion and Future Work}

Though formal verification of different fairness metrics of an ML algorithm for different datasets is an important question, existing verifiers are not scalable, accurate, and extendable to non-Boolean attributes. We propose a stochastic SAT-based approach, {\framework}, that formally verifies independence and separation metrics of fairness for different classifiers and distributions for compound protected groups.
Experimental evaluations demonstrate that {\framework} achieves \textit{higher accuracy} and \textit{scalability} in comparison to the state-of-the-art verifiers, FairSquare and VeriFair, while yielding \textit{higher robustness} than the sample-based tools, such as AIF360.

Our work opens up several new directions of research. One direction is to develop SSAT models and verifiers for popular classifiers like Deep networks and SVMs. Other direction is to develop SSAT solvers that can accommodate continuous variables and conditional probabilities by design.

\section*{Acknowledgments}
We are grateful to  Jie-Hong Roland Jiang and Teodora Baluta for the useful discussion at the earlier stage of this project. We thank Nian-Ze Lee for the technical support of the SSAT solvers. This work was supported in part by the National Research Foundation Singapore under its NRF Fellowship Programme [NRF- NRFFAI1-2019-0004] and the AI Singapore Programme [AISG-RP-2018-005], and NUS ODPRT Grant [R-252-000-685-13]. The computational work for this article was performed on resources of  Max Planck Institute for Software Systems, Germany  and the National Supercomputing Centre, Singapore. 
Debabrota Basu was funded by WASP-NTU grant of the Knut and Alice Wallenberg Foundation during the initial phase of this work.

\clearpage
\bibliography{main} 
\clearpage
\clearpage
\appendix
\section{Proofs of Theoretical Results}

%\begin{lemmarep}
%	Solving the ER-SSAT and RE-SSAT problems are $\mathrm{NP}^{\mathrm{PP}}$ hard~\cite{littman2001stochastic}.
%\end{lemmarep}
%\begin{proof}[Proof of Lemma~\ref{thm:complexity}]
%	The decision version of ER-SSAT problem is 
%	\begin{equation}
%		\Phi := \exists a_1,\dots, \exists a_n, \R^{p_{x_1}}x_1, \dots, \R^{p_{x_m}}x_m. \; \Pr[\phi_{\hat{y}}] \geq t,\notag
%	\end{equation}
%	where $t$ is a threshold in $[0,1]$.
%	It is exactly the an E-MAJSAT (or threshold SAT) problem which is $NP^{PP}$ hard~\cite{littman2001stochastic}.
%	If there's no random variable and $t=1$, ER-SSAT reduces to a SAT problem, which is NP-hard. If there's no existential variable, ER-SSAT reduces to a MAJSAT problem, which is PP-hard.
%	Similar arguments also hold for RE-SSAT problem.
%\end{proof}
%

\begin{lemmarep}
	Given Eq.~\eqref{eq:ar} and~\eqref{eq:er_complement},	$ \Pr[\Phi_{\mathsf{UR}}] = 1 - \Pr[\Phi'_{\mathsf{ER}}]  $.
\end{lemmarep}
\begin{proof}[Proof of Lemma~\ref{thm:dual}]
	Both $ \Phi_{\mathsf{UR}} $ and $ \Phi'_{\mathsf{ER}} $ have  random quantified variables in the identical order in the prefix. According to the definition of SSAT formulas,
\begin{equation}
\Pr[\Phi_{\mathsf{UR}}] = \min\limits_{a_1, \dots,a_n} \Pr[\phi_{\hat{Y}}] \text{ and } \Pr[\Phi'_{\mathsf{ER}}] = \max\limits_{a_1, \dots,a_n} \Pr[\neg\phi_{\hat{Y}}].\notag
\end{equation}
We can show the following duality between ER-SSAT and UR-SSAT,
\begin{equation}
\begin{split}
\Pr[\Phi'_{\mathsf{ER}}] &= \max\limits_{a_1, \dots,a_n} \Pr[\neg\phi_{\hat{Y}}]  \\
&= \min\limits_{a_1, \dots,a_n} (1 - \Pr[\phi_{\hat{Y}}])\\
&= 1 - \min\limits_{a_1, \dots,a_n}  \Pr[\phi_{\hat{Y}}]\\
&= 1 - \Pr[\Phi_{\mathsf{UR}}].\notag
\end{split}
\end{equation}
\end{proof}

\begin{lemmarep}
	Let $ \Phi_{\mathbf{a}} $ be the RE-SSAT formula for computing the PPV of the compound protected group $ \mathbf{a} \in A $. If $ \Phi_{\mathsf{ER}} $ is the ER-SSAT formula for learning the most favored group and $ \Phi_{\mathsf{UR}} $ is the UR-SSAT formula for learning the least favored group, then
	$\max_{\mathbf{a}} \; \Pr[\Phi_{\mathbf{a}}] = \Pr[\Phi_{\mathsf{ER}}]$   
	and
	$\min_{\mathbf{a}} \; \Pr[\Phi_{\mathbf{a}}] = \Pr[\Phi_{\mathsf{UR}}]$.   
\end{lemmarep}
\begin{proof}[Proof of Lemma~\ref{lm:equivalence}] 
	It is trivial that the PPV of most favored group $ \mathbf{a}_{\mathsf{fav}} $ is the maximum PPV of all compound groups $ \mathbf{a} \in A $. Similarly, the PPV of the least favored group $ \mathbf{a}_{\mathsf{unfav}} $ is the minimum PPV of all compound groups $ \mathbf{a} \in A $.
	
	By construction of the SSAT formulas, the PPV of $ \mathbf{a}_{\mathsf{fav}} $ and $ \mathbf{a}_{\mathsf{unfav}} $ are $ \Pr[\Phi_{\mathsf{ER}}] $ and $ \Pr[\Phi_{\mathsf{UR}}] $ respectively. Since $ \Pr[\Phi_{\mathbf{a}}] $ is the PPV of the compound group $ \mathbf{a} $, 
	\begin{equation}
	\begin{split}
		\max_{\mathbf{a}} \; \Pr[\Phi_{\mathbf{a}}] = \Pr[\Phi_{\mathsf{ER}}]
	\text{ and }
	\min_{\mathbf{a}} \; \Pr[\Phi_{\mathbf{a}}] = \Pr[\Phi_{\mathsf{UR}}].\notag
	\end{split}
	\end{equation}
\end{proof}

\begin{theoremrep}
	For an ER-SSAT problem, the sample complexity is given by 
	\[ k = O\left((n+ \ln(1/\delta))\frac{\ln m}{\ln \epsilon_0} \right),\]
	where $\frac{\hat{p}}{p} \leq \epsilon_0$ with probability $1-\delta$ such that $\epsilon_0 \geq 1$.
\end{theoremrep}
%Theorem~\ref{thm:sample} states that for the prescribed number of samples the estimated disparate impact $\hat{DI}$ and statistical parity $\hat{SP}$ would also satisfy $\hat{DI} \leq  \epsilon_0 DI, $ and $\hat{SP} \leq 2\epsilon_0 SP$.
%Apply Hoeffding inequality for each term and then use Union bound for $2^n$ existential variables.
\begin{corollaryrep}
	If $k$ samples are considered from the data-generating distribution in {\framework} such that 
	\[
	k = O\left((n+ \ln(1/\delta))\frac{\ln m}{\ln \epsilon_0}\right),
	\]
	the estimated disparate impact $\hat{DI}$ and statistical parity $\hat{SP}$ satisfy, with probability $1-\delta$,
	$
	\hat{DI} \leq  \epsilon_0 DI, \quad \text{and} \quad \hat{SP} \leq \epsilon_0 SP.
	$
\end{corollaryrep}
\begin{proof}[Proof of Corollary~\ref{cor:error}] 
	By Theorem~\ref{thm:sample}, we get that for $k$ samples obtained from the data generating distribution, where
	$$ k \geq (n+ \ln(1/\delta))\frac{\ln m}{\ln \epsilon_0},$$
	the estimated probability of satisfaction for the most and least favoured groups $\hat{p}_{max}$ and  $\hat{p}_{min}$ satisfies
	\begin{equation}
	\hat{p}_{max} \leq \epsilon_0 \max_{\mathbf{a}} \; \Pr[\Phi_{\mathbf{a}}]
	\text{ and }
	\hat{p}_{min} \leq \epsilon_0 \min_{\mathbf{a}} \; \Pr[\Phi_{\mathbf{a}}].\notag
	\end{equation}
	with probability $1-\delta$.
	Thus, the estimated value of disparate impact will satisfy
	$$\hat{DI} \triangleq \frac{\hat{p}_{max}}{\hat{p}_{min}} \leq \epsilon_0  \frac{p_{max}}{p_{min}} \leq \epsilon_0 DI,$$
	and statistical parity will satisfy 
	$$\hat{SP} \triangleq \abs{\hat{p}_{max} - \hat{p}_{min}} \leq  \epsilon_0  \abs{p_{max} - p_{min}} \leq \epsilon_0 SP,$$
	with probability $1-\delta$.
\end{proof}

\section{Additional Experimental Details}
\subsection{Experimental Setup}
\label{sec:exp_extra}
Since both {\framework}  and FairSquare take a  probability distribution of the attributes as input, we perform five-fold cross validation, use the train set for learning the classifier, compute distribution on the test set and finally verify fairness metrics such as disparate impact and statistical parity difference on the distribution.

\begin{table*}[t!]        
	\centering
	\caption{Verification of different fairness enhancing algorithms for multiple datasets and classifiers using {\framework}. Numbers in bold refer to fairness improvement  compared against the unprocessed (orig.) dataset. RW and OP refer to reweighing and optimized-preprocessing algorithm respectively.}\label{tab:fair_algo_verification_appendix}
	\vspace*{-.2em}
	\begin{tabular}{ll
			ccc
			ccc
			%            				ccc
			%            				ccc
			ccc
			ccc}
		\toprule
		\multirow{3}{*}{Classifier}& Dataset $ \rightarrow $   & 
		\multicolumn{6}{c}{German} \\
		\cmidrule(lr){3-8}
		& Protected  $ \rightarrow $ & 
                \multicolumn{3}{c}{Age}   & \multicolumn{3}{c}{Sex}  
		\\ 
		\cmidrule(lr){3-5}
		\cmidrule(lr){6-8}
		
		& Algorithm  $ \rightarrow $ &  
		orig. & RW & OP &
		orig. & RW & OP \\ 
		\midrule

		\multirow{3}{*}{\shortstack{Logistic \\ regression}}
		& Disparte impact&  $ 0.00 $ &  $ 0.00 $ &  $ \mathbf{0.31} $ &  $ 0.27 $ &  $ \mathbf{0.46} $ &  $ 0.17 $  \\
		& Stat. parity&  $ 0.45 $ &  $ \mathbf{0.03} $ &  $ \mathbf{0.12} $ &  $ 0.03 $ &  $ \mathbf{0.02} $ &  $ 0.07 $  \\
		& Equalized odds&  $ 0.65 $ &  $ \mathbf{0.04} $ &  $ \mathbf{0.14} $ &  $ 0.10 $ &  $ \mathbf{0.08} $ &  $ 0.13 $  \\
		\midrule
		\multirow{3}{*}{\shortstack{Decision \\ tree}}
		& Disparte impact&  $ 0.00 $ &  $ \mathbf{0.56} $ &  $ \mathbf{0.12} $ &  $ 0.35 $ &  $ \mathbf{0.37} $ &  $ \mathbf{0.38} $  \\
		& Stat. parity&  $ 0.35 $ &  $ \mathbf{0.02} $ &  $ \mathbf{0.22} $ &  $ 0.05 $ &  $ 0.10 $ &  $ 0.11 $  \\
		& Equalized odds&  $ 0.36 $ &  $ \mathbf{0.05} $ &  $ \mathbf{0.28} $ &  $ 0.06 $ &  $ 0.16 $ &  $ 0.17 $  \\

		\bottomrule
	\end{tabular}
\end{table*}

\end{document}